\documentclass[lettersize,journal]{IEEEtran}
\usepackage{amsmath,amsfonts}
\usepackage{algorithm}
\usepackage{array}
\usepackage[caption=false,font=normalsize,labelfont=sf,textfont=sf]{subfig}
\usepackage{textcomp}
\usepackage{stfloats}
\usepackage{url}
\usepackage{verbatim}
\usepackage{graphicx}

\usepackage{anyfontsize}
\fontsize{9.5pt}{11.5pt}\selectfont

\usepackage{longtable}
\usepackage{cuted}
\pdfoutput=1
\usepackage{amsmath}
\usepackage{amsthm}
\usepackage{amssymb}
\usepackage{natbib}
\usepackage{stfloats}

\usepackage{graphicx}%
\usepackage{multicol}
\usepackage{amsmath,amssymb,amsfonts}%

\usepackage{mathrsfs}%
\usepackage{textcomp}%
\usepackage{manyfoot}%
\usepackage{booktabs}%
\usepackage{algorithmicx}%
\usepackage{algpseudocode}%
\usepackage{listings}%
\usepackage{tcolorbox}
\usepackage{epigraph}
\usepackage{bbding}
\usepackage{multirow}
\usepackage{rotating}
\usepackage{xcolor}
\usepackage{tabularx}
\usepackage{enumitem}
\usepackage[framemethod=TikZ]{mdframed}
\usepackage[parfill]{parskip}
\usepackage[switch]{lineno}
\definecolor{lightgray}{gray}{0.9}
\definecolor{topology}{RGB}{157,31,99}
\definecolor{geometry}{RGB}{240,90,19}
\definecolor{algebra}{RGB}{0,77,244}

\usepackage[utf8]{inputenc}

\newlist{DataPoint}{enumerate}{1}
\setlist[DataPoint]{label={C\arabic*:},leftmargin=*,itemindent=0em}
\newlist{DataSignal}{enumerate}{1}
\setlist[DataSignal]{label={S\arabic*:},leftmargin=*}

\newcommand{\nonindentedDataPoint}[1]{%
  \end{DataPoint} 
  \noindent#1 
  \begin{DataPoint}[resume] 
}

\definecolor{shadecolor}{RGB}{242,242,242}
\mdfdefinestyle{tcolorbox}{
  backgroundcolor=shadecolor,
  linecolor=darkgray,
  linewidth=1.5pt,
  roundcorner=5pt,
  nobreak=true,
  innerleftmargin=15pt,
  innerrightmargin=15pt,
  innertopmargin=10pt,
  innerbottommargin=10pt,
  skipabove=10pt,
  skipbelow=10pt,
}

\usepackage{hyperref}
\hypersetup{
	colorlinks=true,
	breaklinks=true,
    citecolor=black,
	urlcolor=black, 
	linkcolor=black,
}

\usepackage{soul} 

\makeatletter
\newcommand*{\citelinktext}[2]{%
  \hyper@@link[cite]{}{cite.#1}{#2}}
\makeatother

\begin{document}

\title{\vspace{1cm}
 Beyond Euclid: \\ \Large An Illustrated Guide to Modern Machine Learning \\ with Geometric, Topological, and Algebraic Structures}

\author{
  \IEEEauthorblockN{
    Mathilde Papillon$^{*1 \dagger}$
    Sophia Sanborn$^{*1,2}$
    Johan Mathe$^{*3}$
    Louisa Cornelis$^{*1}$
    Abby Bertics$^1$
    Domas Buracas$^4$\\
    Hansen J Lillemark$^{4,6}$
    Christian Shewmake$^{4}$
    Fatih Dinc$^{1}$
    Xavier Pennec$^5$
    Nina Miolane$^{1, 3, 4}$
  } \\ \vspace{1em}
  \small{
    \IEEEauthorblockA{$^1$UC Santa Barbara} \quad
    \IEEEauthorblockA{$^2$Stanford University} \quad
  \IEEEauthorblockA{$^3$Atmo, Inc.} \quad
  \IEEEauthorblockA{$^4$New Theory AI}} \quad
  \IEEEauthorblockA{$^5$Université Côte d’Azur \& Inria} \quad
  \IEEEauthorblockA{$^6$UC Berkeley} \quad \\
  \vspace{0.15cm}
  \small{$^\dagger$ Corresponding author: \texttt{papillon@ucsb.edu}}
  \vspace{-0.9cm}
}

\maketitle

\begin{abstract}
The enduring legacy of Euclidean geometry underpins classical machine learning, which, for decades, has been primarily developed for data lying in Euclidean space. Yet, modern machine learning increasingly encounters richly structured data that is inherently non-Euclidean. This data can exhibit intricate geometric, topological and algebraic structure: from the geometry of the curvature of space-time, to topologically complex interactions between neurons in the brain, to the algebraic transformations describing symmetries of physical systems. Extracting knowledge from such non-Euclidean data necessitates a broader mathematical perspective. Echoing the 19th-century revolutions that gave rise to non-Euclidean geometry, an emerging line of research is redefining modern machine learning with non-Euclidean structures. Its goal: generalizing classical methods to unconventional data types with geometry, topology, and algebra. In this review, we provide an accessible gateway to this fast-growing field and propose a graphical taxonomy that integrates recent advances into an intuitive unified framework. We subsequently extract insights into current challenges and highlight exciting opportunities for future development in this field.

\end{abstract}



\section{Introduction}\label{sec:introduction}

For nearly two millennia, Euclid's \textit{Elements of Geometry} formed the backbone of our understanding of space and shape. This `Euclidean' view of geometry\textemdash characterized by flat planes and straight lines\textemdash remained unquestioned until the 19th century. Only then, did mathematicians venture ``beyond'' to develop the principles of non-Euclidean geometry on curved spaces. Their pioneering work revealed that there is no singular geometry. Instead, Euclidean geometry is but one in a mathematical universe of geometries, each of which can be used to illuminate different structures in nature\textemdash from the mechanics of celestial bodies embracing the curvature of spacetime to the topologically and algebraically complex electrical patterns of neurons in natural and artificial neural networks.

This non-Euclidean revolution was part of a greater trend towards generalization and abstraction in 19th and 20th century mathematics. In addition to expanding the realm of \textit{geometry}, mathematicians proceeded to define more abstract notions of space, freed from rigid geometric concepts like distances and angles. This gave rise to the field of \textit{topology}, which examines the properties of a space that are preserved under continuous transformations such as stretching and bending. By abstracting away from the rigidity of geometric structures, topology emphasizes more general spatial properties such as continuity and connectedness. Indeed, two structures that look very different from a geometric perspective may be considered topologically equivalent. The famous of example of this is a donut and coffee mug, which are topologically equivalent since one can be continuously deformed into the other. This notion of abstract equivalence was supported by the simultaneous development of the field of abstract \textit{algebra}, which examines the symmetries of an object\textemdash the transformations that leave its fundamental structure unchanged. These mathematical ideas quickly found applications in the natural sciences, and revolutionized how we model the world.

A similar revolution is now unfolding in machine learning, \textcolor{black}{(see for example \citet{Bronstein2017beyond})}. In the last two decades, a burgeoning body of research has expanded the horizons of machine learning, moving beyond the flat, Euclidean spaces traditionally used in data analysis to embrace the rich variety of structures offered by non-Euclidean geometry, topology, and abstract algebra. This movement includes the generalization of classical statistical theory and machine learning in the field of \textit{Geometric Statistics} \citep{Pennec2006,guigui2023introduction} as well as deep learning models in the fields of \textit{Geometric}, \citep{bronstein2021geometric}, \textit{Topological} \citep{hajijtopological,bodnar2023topological}, and \textit{Equivariant} \citep{cohen2021equivariant} \textit{Deep Learning}. In the 20th century, non-Euclidean geometry radically transformed how we model the world with pen and paper. In the 21st century, it is poised to revolutionize how we model the world with machines.

This review article provides an accessible introduction to the core concepts underlying this movement. We organize the models in this body of literature into a coherent taxonomy defined by the mathematical structure of both the \textit{data} and the machine learning \textit{model}. In so doing, we clarify distinctions between approaches and we highlight challenges and high-potential areas of research that are as-of-yet unexplored. We begin by introducing the essential mathematical background in Section~\ref{sec:elements}, and turn to an analysis of mathematical structure in data in Section~\ref{sec:data} before introducing our ontology of machine learning and deep learning methods in Section \ref{sec:ml} and \ref{sec:dl}. We explore the associated landscape of open-source software libraries in Section~\ref{sec:software}, and delve into the movement's key application domains in Section~\ref{sec:apps}. Accordingly, this review article reveals how machine learning born from the elegant mathematics of geometry, topology, and algebra has been developed, implemented, and adapted to propose transformative solutions to real-world challenges.

\begin{figure*}[!ht]
    \centering
    \includegraphics[width=0.85\linewidth]{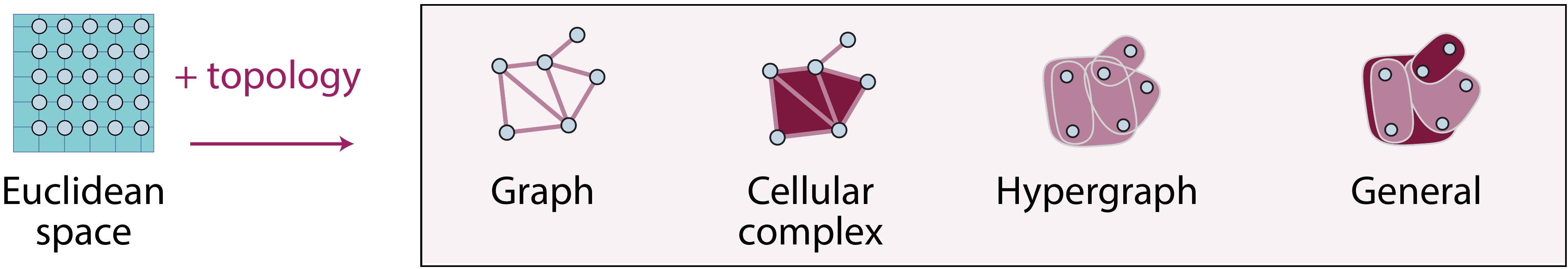}
    \caption{\textbf{Beyond Euclid: Discrete Topological Structures.} Left: Euclidean space discretized into a regular grid. Right: Discrete topological spaces that go beyond classical discretized Euclidean space. Graphs, Cellular Complexes, Hypergraphs relax the assumption of the regular grid and allow points to be connected with more complex relationships. The arrow \texttt{+topology} indicates the addition of a non-Euclidean, discrete topological structure. Adapted from \cite{papillon2023architectures}.}
    \label{fig:elements_topology}
\end{figure*}

\begin{figure*}[!ht]
    \centering
    \includegraphics[width=0.85\linewidth] {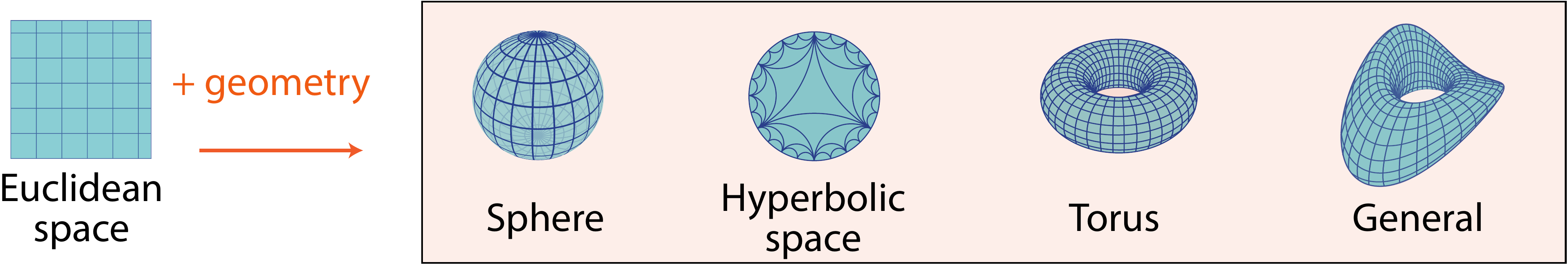}
    \caption{\textbf{Beyond Euclid: Continuous Geometric Structures}. Left: Euclidean space. Right: Riemannian manifolds that go beyond the classical Euclidean space. Spheres, hyperbolic spaces, and tori relax the assumption of flatness of the Euclidean space and can exhibit positive or negative curvature. The arrow \texttt{+geometry} indicates the addition of a non-Euclidean, continuous geometric structure.}
    \label{fig:elements_geometry}
\end{figure*}

\section{Elements of Non-Euclidean Geometry}
\label{sec:elements}

We first provide an accessible and concise introduction to the essential mathematical concepts. For readability, we define the concepts primarily linguistically here, and refer the reader to the works \cite{guigui2023introduction,bronstein2021geometric,hajijtopological,cohen2021equivariant} for their precise mathematical definitions.

\textit{Topology} (shown throughout with \textcolor{topology}{purple terms}), \textit{geometry} (shown throughout with \textcolor{geometry}{orange terms}), and \textit{algebra} (shown throughout with \textcolor{algebra}{blue terms}), are branches of mathematics that study the properties of abstract spaces. 
In machine learning, data can possess explicit spatial structure\textemdash such as an image of a brain scan, or a rendering of a protein surface. Even when data is not overtly spatial, a dataset can be naturally conceptualized as a set of samples drawn from an abstract surface embedded in a high-dimensional space. Understanding the ``shape'' of data\textemdash that is, the shape of the space to which this data belongs\textemdash  can give important insights into the patterns of relationships that give data its meaning. 

Topology, geometry and algebra each provide a different lens and set of tools for studying the properties of data spaces and their ``shapes.'' 
\textbf{Topology} lends the most abstract, flexible perspective and considers spaces as stretchy structures that can be continuously deformed so long as \textit{connectivity} and \textit{continuity} are preserved. Topology thus studies the \textit{relationships} between points. \textbf{Geometry} allows us to quantify familiar properties such as \textit{distances} and \textit{angles}, in other words: perform \textit{measurements} on points. \textbf{Algebra} provides the tools to study the \textit{symmetries} of an object\textemdash the transformations that can be applied while leaving its fundamental structure invariant.

\subsection{\textcolor{topology}{Topology: Relationships}}

A \textbf{topological space} is a set of points equipped with a structure known as a \textit{topology} that establishes which points in the set are ``close'' to each other. The topology gives spatial structure to an otherwise unstructured set. Formally, a topology is defined as a collection of \textbf{open sets}. An open set is a collection of points in a spatial region that excludes points on the boundary. {\color{black}By grouping points into open sets, we can use terms like 'neighborhoods' and 'paths' to reference ``closeness'' and other abstract relationships between points. This gives us a way to formalize concepts like continuity (can one travel from one location to another without teleporting?) and connectedness (are two locations in the same neighborhood or region?).}

Given the generality of topological structures, topological spaces can be quite exotic. 
Within this paper, we do not consider continuous topological spaces, and we explicitly restrict the term \textbf{topology} to \textbf{discrete topological structures}, such as graphs, cellular complexes, and hypergraphs. We consider these spaces to be a generalization of the discretized Euclidean space. Indeed, Euclidean space discretizes into a regular grid, while graphs, cellular complexes, and hypergraphs \textcolor{black}{allow for more flexible patterns of connectivity, where points may interact through complex relationships, as shown in Figure \ref{fig:elements_topology}.}
\begin{figure*}[!ht]
    \centering
    \includegraphics[width=0.85\linewidth]{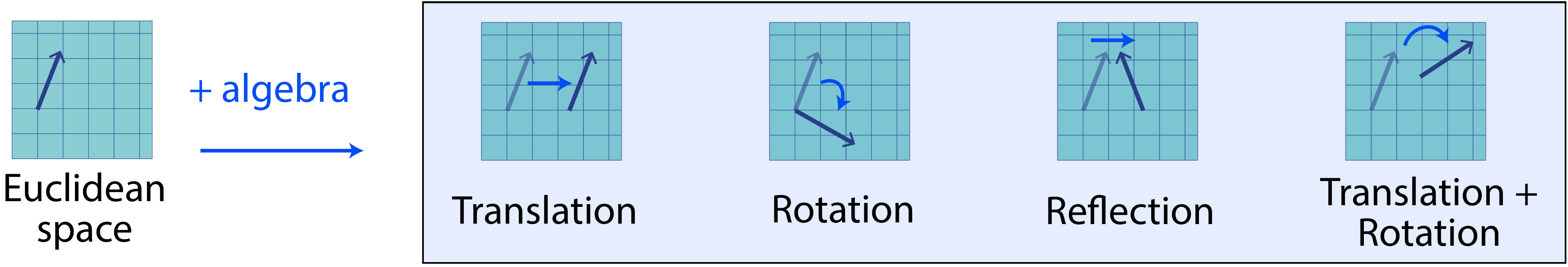}
    \caption{\textbf{Beyond Euclid: Algebraic Transformations}. Left: Euclidean space. Right: Group transformations that act on the elements of an Euclidean space: 2D translation from the group $\mathbb{R}^2$, 2D rotation from the group $SO(2)$, 2D reflection from the group $\{1, -1\}$, and a combination of translation and rotation from the Special Euclidean group $SE(2)$. The arrow \texttt{+algebra} indicates the addition of the non-Euclidean algebraic structure defining a group action. \vspace{-0.2cm}}
    \label{fig:elements_algebra}
\end{figure*}

\subsection{\textcolor{geometry}{Geometry: Measurements}}

A \textbf{manifold} is a continuous space that locally ``resembles'' (is homeomorphic to) Euclidean space in the neighborhood of every point. In other words, it is locally linear, and it does not intersect itself. Though locally it resembles flat space, its global shape may exhibit curvature. Without additional structure, the manifold can be seen as a soft and elastic surface. \textcolor{black}{Introducing a notion of \textit{distance} gives the manifold a more definite structure, as enforcing how far apart points are constrains its overall geometry.}

There are several approaches to defining a distance on the manifold. \textcolor{black}{A powerful method is to use a 
\textbf{Riemannian manifold}, which can be thought of as a smooth high-dimensional surface that locally resembles flat Euclidean space but may curve globally. 
To measure distance on such a surface, we use a 
\textbf{Riemannian metric}. Intuitively, this acts like a local ruler at each point, telling us how to measure lengths of curves that pass through that point. Mathematically, it is defined as \textcolor{black}{a positive definite inner product that varies smoothly on the tangent space at each point}. By integrating these local measurements along a curve, we can compute the total distance along that curve.}
A \textbf{geodesic} generalizes the Euclidean concept of a ``straight line'' to curved Riemannian manifolds. A Riemannian geodesic is a curve that traces the locally shortest path between two points.

There exist different flavors of Riemannian manifolds, such as the spheres, hyperbolic spaces, and tori. We consider these spaces to be generalizations of the continuous Euclidean space. Indeed, Euclidean spaces are globally flat, while spheres, hyperbolic spaces and tori can exhibit curvature \textemdash as illustrated in Figure~\ref{fig:elements_geometry}. 

\vspace{-0.2cm}
\subsection{\textcolor{algebra}{Algebra: Transformations}}

\textcolor{black}{A \textbf{group} is a set of elements equipped with a rule for combining them, called a binary operation. This operation must satisfy certain properties: closure, associativity, the existence of an identity element, and the existence of inverses. While a group itself is just a set with these properties, in many contexts, such as ours, the elements of the group can be interpreted as transformations, like 3D rotations. This interpretation arises when we define a \textbf{group action} or \textbf{representation}, where the elements of the group act as operations on a space.} Groups can be discrete or continuous. A \textbf{Lie group} is a continuous group, defined as a smooth manifold equipped with a compatible group structure such that the composition of elements on the manifold obeys the group axioms. An example of a Lie group is the set of special orthogonal matrices in $\mathbb{R}^{3\times 3}$ under matrix multiplication, which defines the group of 3-dimensional rotations, $SO(3)$. Each matrix is an element of the group and defines rotation at a certain angle. Lie groups are extensively used in physics where they describe symmetries in physical systems.

A group of transformations, such as $SO(3)$\textemdash the group of 3D rotations\textemdash may act on another manifold to transform its elements. A \textbf{group action} maps an element in a manifold to a new location, determined by the group element that transforms them. For example, a group action on a Euclidean space can translate, rotate and reflect its elements, as illustrated in Figure~\ref{fig:elements_algebra}. 
In this paper, we use the term \textbf{algebra} to denote that we equip a space with a \textbf{group action}. {\footnote{\color{black}We recognize that ``algebra'' can have a more general meaning in other contexts, such as in categorical deep learning, where it refers to a structure map relative to an endofunctor within a category. However, within this review, we restrict its meaning to the definition provided above.}}

\section{Structure in Data}\label{sec:data}
\begin{figure*}[ht]
    \centering
    \includegraphics[width=0.9\linewidth]{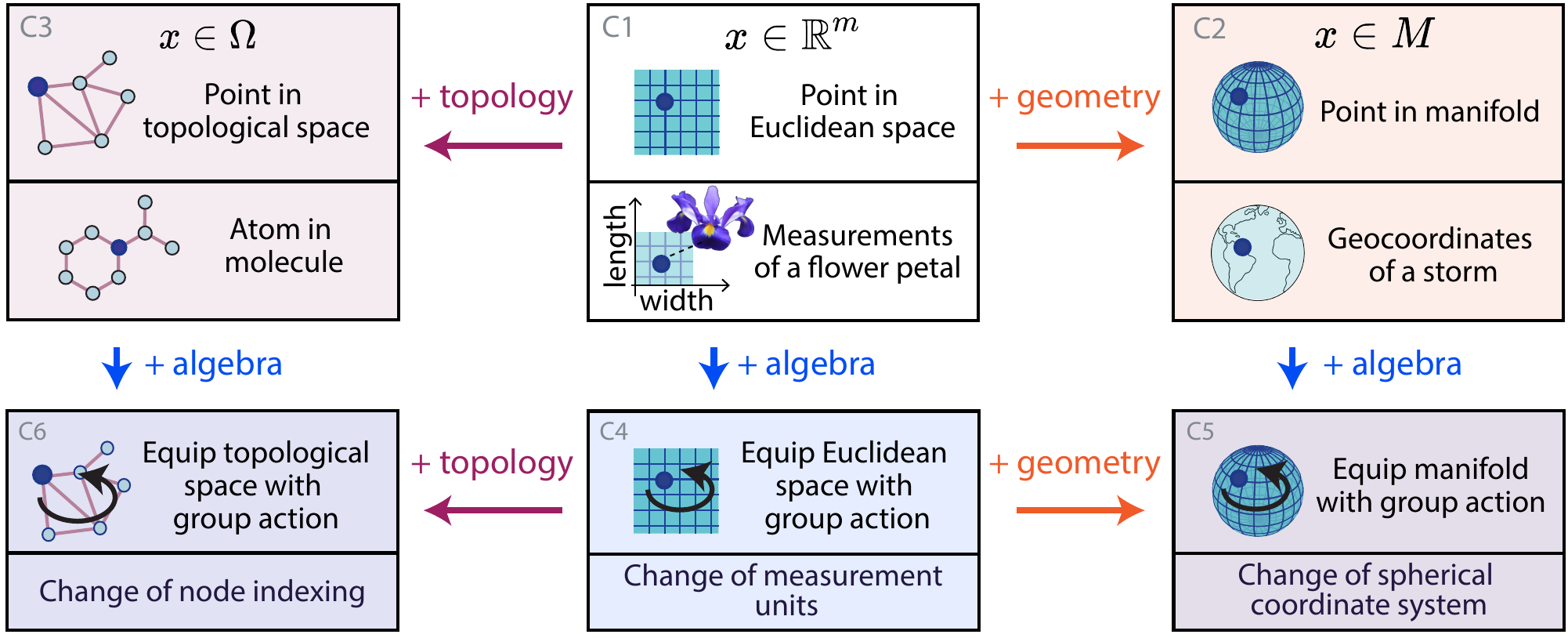}
    \caption{\textbf{Geometric, Topological, and Algebraic Structures in Data as Coordinates.} Each card illustrates and exemplifies a type of coordinate. The arrows \texttt{+ topology}, \texttt{+ geometry} and \texttt{+ algebra} between cards indicate the addition of non-Euclidean topological, geometric and algebraic structures respectively. Notations: $\mathbb{R}^m$: Euclidean space of dimension $m$, $M$: Manifold, $\Omega$: Topological space; $x$: data as point or as signal. C1 flower image credit: Oleg Yunakov / CC BY-SA 4.0 \citep{yunakov2011iris} . \vspace{-0.5cm}}
    \label{fig:data-coord}
\end{figure*}

The mathematics of topology, geometry and algebra provide a conceptual framework to categorize the nature of data found in machine learning. We generally encounter two types of data: either data as \textit{coordinates in space} (Fig. \ref{fig:data-coord}){\color{black}\textemdash for example, the coordinate of the position of an object in a 2D space; or data as \textit{signals over a space} (Fig. \ref{fig:data})\textemdash for example, an image seen as a 3D (RGB) signal defined over a 2D space (the spatial location).} In each case, the \textit{space} can either be a Euclidean space or it can be equipped with topological, geometric and algebraic structures such as the ones introduced in Section~\ref{sec:elements}.

Understanding the structure of this space provides essential insights into the nature of the data. These insights \textcolor{black}{can be}, in turn, crucial for selecting the machine learning model that will be most suitable to extract knowledge from this data.  \textcolor{black}{We note the emergence of a line of research which leverages Euclidean architectures for dealing with non-Euclidean data, such as \textcolor{black}{protein folding} \citep{abramson2024accurate}, and refer the reader to \citet{brehmer2024does} for a discussion on the topic.} Next, we provide a graphical taxonomy, shown in Figure~\ref{fig:data}, to categorize the structures of data based on the mathematics of topology, geometry and algebra.

\subsection{Data as Coordinates in Space}

We start by categorizing the mathematical structures of data, when data points are \textit{coordinates in space}\textemdash as shown in Figure~\ref{fig:data-coord}, and described in detail below. We denote $x$ \textcolor{black}{as a datapoint} within a dataset $x_1, ..x_i, ..., x_N$, and drop its subscript $i$ for \textcolor{black}{conciseness.}

\begin{DataPoint}
    \item \textbf{Point in Euclidean space}: This category of data points forms the basis of conventional machine learning and deep learning approaches. Card C1 (white) in Figure~\ref{fig:data} shows an example of a point in Euclidean space \textcolor{black}{$\mathbb{R}^m$}, which represents the dimensions of a flower from the Iris dataset \citep{fisher36lda}, where $n$ represents the number of features studied.
    \item \textbf{\textcolor{geometry}{Point in manifold}}: Adding a non-Euclidean geometry to the coordinate space, card C2 (orange) illustrates a data point on a manifold $M$, where the manifold is the sphere. For example, this data point can represent the geographic coordinates of a storm event, i.e., its location on the surface of the earth represented as a sphere. 
    \item \textbf{\textcolor{topology}{Point in topological space}}: Card C3 \textcolor{black}{(light purple)} considers a data point that resides in a topological space $\Omega$, here corresponding to a node in a graph. {\color{black}An example in this category would be a data point representing an atom in the graph of a molecule, in which edges represent the bounds between the atoms.}

    \nonindentedDataPoint{{{The next three data categories add a group action to the data spaces mentioned above.}}}

    \item \textbf{Point in Euclidean space \textcolor{algebra}{equipped with group action}}: Card C4 (blue) displays a data point in a Euclidean space that has been equipped with a group action. Group actions enable us to model and preserve symmetries in the data. For example, a group action on the flower dimensions could be the change of units, from centimeters to millimeters. In this case, the group of interest is the group of scalings $\mathbb{R}^*_+$. The action of this group does not change the information contained in the data (the flower will still have the same size in the real world), it only changes the way we encode that information. 
    \item \textbf{\textcolor{geometry}{Point in manifold} \textcolor{algebra}{equipped with group action}}: Card C5 adds the notion of group action \textcolor{black}{to a manifold}. An example of such a group action could be a change of coordinate systems on the sphere, changing the origin of longitudes. In this case, the group of interest would be the group of 3D rotations $SO(3)$. Again, the action of this group represents symmetries in the data: the information content is unchanged (a storm will still happen at the same geographical location in the real world) but the way we encode that information has changed.
    \item \textbf{\textcolor{topology}{Point in topological space} \textcolor{algebra}{equipped with group action}}: Card C6 equips a topological space with the notion of group action. An example of this is a graph of people equipped with an action that can change the way indexing is done on the dataset. The order in which we index people does not change their social relationships, only the way we represent this data in the computer. The group of interest in this case in the group of permutations.
\end{DataPoint}

These categories describe the \textcolor{black}{\textit{mathematical structure}} of the data spaces encountered in machine learning. However, even when data points belong to spaces with topological, geometric, or algebraic structures, their \textcolor{black}{\textit{computational representations}} typically take the form of arrays. These data points may thus appear as vectors in a computer's memory, but this is merely a convenience for processing and storage. The underlying mathematical structure is preserved through constraints imposed on the values of these vectors. 

We note that a data point on a manifold exists independently of the array with which a computer represents. For example, the data point on the sphere can be represented by a vector of size 3 encoding its Cartesian coordinates, or by a vector of size 2 representing its latitude and longitude. In both cases, the mathematical nature of the data point is unchanged: it still represents a point on a manifold. For instance, when represented as a 3D coordinate vector, a data point on the sphere is constrained to have unit norm (i.e., its Euclidean length equals one), which encodes the geometric properties defining the sphere. More generally, a norm defines a notion of length in a vector space, and can induce a metric that measures distances between points. In the context of 3D rotations, \citet{zhou2019continuity} provides an experimental analysis of their computational representations and the impact of representation choice on machine learning models.

\subsection{Data as Signals}

\begin{figure*}
    \centering
    \includegraphics[width=0.9\linewidth]{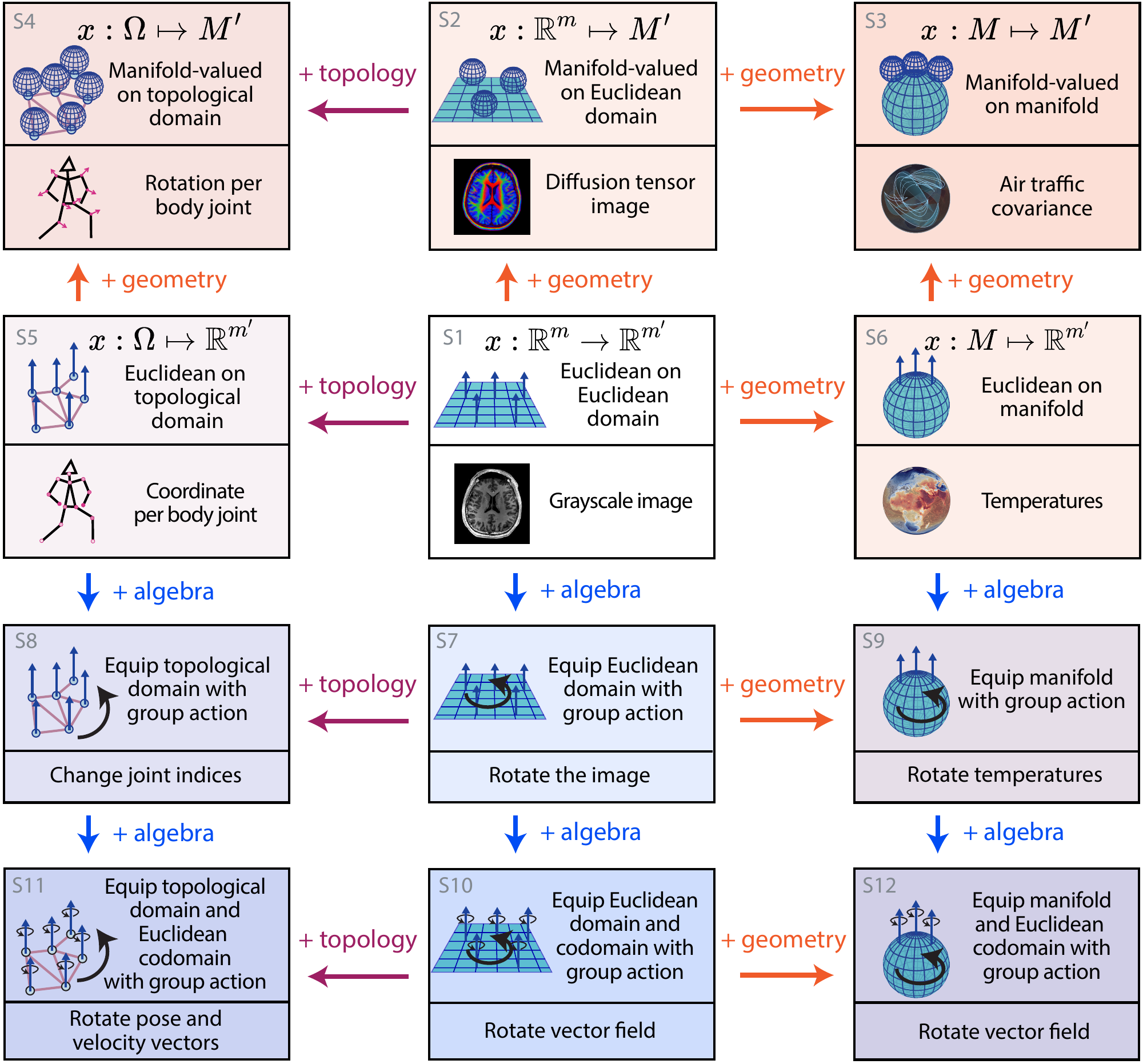}
    \caption{\textbf{Geometric, Topological, and Algebraic Structures in Data as Signals.} Each card illustrates the structure of a signal and presents a real-world example. The arrows \texttt{+ topology}, \texttt{+ geometry} and \texttt{+ algebra} between cards indicate the addition of non-Euclidean topological, geometric and algebraic structures respectively. Notations: $\mathbb{R}^m$: Euclidean space of dimension $m$, $M$: Manifold, $\Omega$: Topological space; $x$: data as point or as signal. S1 image credit: \citet{asnaebsa2022mri} / CC-BY-SA 4.0; S2 image credit: adapted from \citet{asnaebsa2022mri} / CC-BY-SA 4.0; S6 image credit: © Atmo, Inc. — used with permission. \vspace{-0.5cm}}
    \label{fig:data}
\end{figure*}

In many applications, data points are not coordinates in space but rather \textit{functions} defined over a space, typically assigning a vector in $\mathbb{R}^{m'}$ to every point in the space $\mathbb{R}^m$. Formally, we can write such a function as $x: \mathbb{R}^m \rightarrow \mathbb{R}^{m'}$ where the input space (here, $\mathbb{R}^m$) is called the domain and the output space (here, $\mathbb{R}^{m'}$) is called the codomain. We refer to data of this type as a \textbf{signal}. Elements of the codomain are called the values of the signal $x$. 

Color images provide a clear example of this. The domain is a bounded region of $\mathbb{R}^2$: for example, limited to the range $[0, 1920]$ on the horizontal axis and $[0, 1080]$ on the vertical axis such that each point in the domain represents a pixel location. Each location is assigned a vector in the codomain $\mathbb{R}^3$, specifying the intensity on each RGB color channel. 

Figure~\ref{fig:data} introduces several types of data points as functions from a domain to a codomain, together with real-world data examples. Formalizing a signal in this way gives us the flexibility to represent more general classes of signals using non-Euclidean structures. The domain or the codomain (or both) may be any one of the non-Euclidean spaces introduced in the previous section. We present the different options in the four bottom rows of Figure~\ref{fig:data} and detail them below.

\begin{DataSignal}
    \item \textbf{Euclidean signal on Euclidean domain}: This represents the most common type of data in classical deep learning, functions as $x_i: \mathbb{R}^{n} \rightarrow \mathbb{R}^{n^\prime} $, going from points in domain $\mathbb{R}^n$ to features in codomain $\mathbb{R}^{n'}$. A gray-scale image provides a clear example of this. The domain is a bounded region of $\mathbb{R}^2$, which is discretized into integer steps, \textit{i.e.}, pixels. \textcolor{black}{Typically, this domain is computationally modeled as $\mathbb{Z}^2$.} Each pixel location is assigned a vector in codomain $\mathbb{R}$, which specifies the intensity in \textcolor{black}{grayscale}. 
    \item \textbf{\textcolor{geometry}{Manifold-valued signal} on Euclidean domain}: This represents signals that can be formalized as a function $x_i: \mathbb{R}^n \mapsto M'$, which assigns an element of a manifold $M'$ to every point in the space. An example of this occurs in diffusion tensor imaging, in which 3D images of the brain are taken, characterized as voxels in $\mathbb{R}^3$. Attached to each voxel is a $3 \times 3$ covariance matrix that describes how water molecules diffuse in the brain \textemdash i.e., an element from the manifold of symmetric positive definite matrices $SPD(3)$~\citep{Pennec2006}.
    
    \item \textbf{\textcolor{geometry}{Manifold-valued signal on manifold domain}}: In this category, both the point coordinates and the features live on manifolds $M$ and $M'$, respectively, i.e., $x_i: M \mapsto M'$. Air traffic covariance data is a usecase where the domain is the \textcolor{black}{two-sphere} $S^2$ (earth surface) and the features are $2\times S^2$ SPD matrices, i.e., elements of the manifold codomain $SPD(2)$. These matrices can encode covariance matrices representing representing different levels of local complexity \citep{LeBrigant2018}.
    
    \item \textbf{\textcolor{geometry}{Manifold-valued signal} \textcolor{topology}{on topological domain}}: Here, the coordinates of the point live in a topological domain $\Omega$ and the features live on a manifold $M'$, such that $x_i : \Omega \mapsto M'$. A representative example for this setting is \textcolor{black}{a} human pose. Each joint in the body is a feature encoded on the special orthogonal group $SO(3)$, and the domain can be a non-directed graph.
    
    \item \textbf{Euclidean signal \textcolor{topology}{on topological domain}}: This represents a signal where point coordinates live on a topological domain $\Omega$ and the features live in a Euclidean space $\mathbb{R}^{n'}$, that is: $x_i : \Omega \mapsto \mathbb{R}^{n'}$. Reusing the example of encoding the human pose, each joint can have a feature encoded \textcolor{black}{in $\mathbb{R}^{3}$} defined over a non directed graph domain.
    \item \textbf{Euclidean signal \textcolor{geometry}{on manifold domain}}: This represents the coordinates of the points on a manifold $M$ and the features in Euclidean space $\mathbb{R}^{n'}$, i.e., $x_i : M \mapsto \mathbb{R}^{n'}$. An example is a dataset in which each datapoint is a snapshot of the Earth at a given time showing the distribution of temperatures across the globe: the surface temperatures live on the sphere manifold $S^2$ and the features are in $\mathbb{R}$ (image and dataset credits: Atmo, Inc.).
    \item \textbf{Euclidean signal on Euclidean domain \textcolor{algebra}{equipped with domain action}}: Card S7 has function, domain, and features that are the same as in Card S1: $x_i: \mathbb{R}^{n} \rightarrow \mathbb{R}^{n^\prime} $, but the Euclidean domain is now equipped with a group action. For example, the domain $\mathbb{R}^2$ of an image can be equipped \textcolor{black}{with the group action of the wallpaper group $P4$, which enables 90° rotations of the image.}
    \item \textbf{Euclidean signal \textcolor{topology}{on topological domain} \textcolor{algebra}{equipped with domain action}}: The function here is $x_i : \Omega \mapsto \mathbb{R}^{n'}$, where $\Omega$ is equipped with a group action. An example of application here is the pose of the human body, where the domain is a undirected graph representing the body joints and the group action on the domain here is the group of permutation matrices $ P $.
    \item \textbf{Euclidean signal \textcolor{geometry}{on manifold domain} \textcolor{algebra}{equipped with a domain action}}: Here, the function is $x_i : M \mapsto \mathbb{R}^{n'}$, where the manifold domain $M$ is equipped with a group action. Using the earth surface temperature example previously defined, we can apply a rotation on the domain of temperature domain, in the case of a change of spherical coordinates for instance. In that case, the action group is the group of rotations $SO(3)$.
    \item \textbf{Euclidean signal on Euclidean domain \textcolor{algebra}{equipped with domain and codomain actions}}: This represents functions such as $x_i: \mathbb{R}^{n} \rightarrow \mathbb{R}^{n^\prime} $, where both $\mathbb{R}^n$ and $\mathbb{R}^{n'}$ are equipped with group actions. The illustration in the cards shows a vector field in a domain $\mathbb{R}^2$ with features (vectors) in $\mathbb{R}^2$. In this example, the actions on both the domain and the codomain are from the group of rotations $SO(2)$, which applies the same rotation to each point and for each independent vector in the vector field.
    \item \textbf{Euclidean signal \textcolor{topology}{on topological domain} \textcolor{algebra}{equipped with domain and codomain actions}}: Here, the function is $x_i : \Omega \mapsto \mathbb{R}^{n'}$, where both $\Omega$ and $\mathbb{R}^{n'}$ are equipped with group actions. Similarly to the previous example, we can apply a permutation group $P$ on the domain and the action of the rotation group $SO(3)$ on the features. \textcolor{black}{Another relevant example are geometric molecular graphs which also exhibit permutation and Euclidean symmetry.}
    \item \textbf{Euclidean signal \textcolor{geometry}{on manifold domain} \textcolor{algebra}{equipped with domain and codomain actions}}: Here $x_i : M \mapsto \mathbb{R}^{n'}$, where both $M$ and $\mathbb{R}^{n'}$ are equipped with group actions. We present an example of a vector field defined over the sphere $S^2$ with vectors in $\mathbb{R}^{n'}$. Applying a the same group action -- an $SO(3)$ rotation -- rotates both the coordinates of the vectors and their direction. This can be useful for changes of coordinates.
\end{DataSignal}

As in the previous subsection, we emphasize the difference between the \textcolor{black}{\textit{mathematical structure}  of data and the \textit{computational representation}} of data. Until now, we described data as signals represented as functions $x$ over a domain. However, most of the time, the functions $x$ are discretized in the computer. For example, the domain of a function representing a 2D color image as the signal $x: \mathbb{R}^2 \mapsto \mathbb{R}^{3}$ is discretized into a grid with $p^2$ pixels, where each pixel has a value in $\mathbb{R}^{3}$. The data point $x$ is therefore represented, in the computer, by an array of length $p^2*3$. A recent line of research, the literature of operators and neural operators \citep{kovachki2023neural}, offers a different approach by treating each data point $x$ as a continuous function without discretization.

\vspace{0.1cm}

\paragraph*{Remark: Opportunities} The cases of manifold-valued signals on manifold domain, topological-valued signals on topological domains, or topological-valued signals on manifold domains are not included in the table, since they have rarely been considered in the machine learning and deep learning literatures. These classes represent an avenue for future research.

\vspace{0.1cm}

\paragraph*{Remark: Data as Spaces} Beyond data as coordinates and data as signals, a third class can be considered: \textit{data as spaces}, where a data point $x$ \textit{is} itself a manifold or topological space. For example, in a molecular dataset, each molecule $x$ can be represented as a graph capturing its structure, with atoms as nodes and bonds as edges; the dataset is then a collection of graphs. Similarly, in datasets of 3D scans, such as heart surfaces, each $x$ is a distinct manifold. However, in most cases, $x$ carries additional structure beyond its topology. A molecule, for instance, is not only defined by atomic bonds but also by the 3D positions of its atoms—thus represented as a signal from a graph $\Omega$ to $\mathbb{R}^3$, as in Card S5. Likewise, heart surfaces are described by 3D coordinates at each point, making each $x$ a signal from a manifold $M$ to $\mathbb{R}^3$. Consequently, these examples ultimately fall into the category of data as signals.

This review uses the two main classes of data representations \textbf{as coordinates}, and \textbf{as signals}, to describe non-Euclidean structure in machine learning and deep learning.

\begin{figure}
    \centering
    \includegraphics[width=0.65\linewidth]{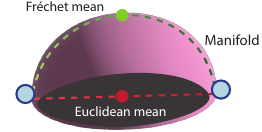}
    \caption{{\color{black}The Fréchet mean lies on the manifold, unlike the Euclidean mean.}}
    \label{fig:frechet}
\end{figure}

\section{Survey: Non-Euclidean Machine Learning}\label{sec:ml}

We now review a large and disparate body of literature of non-Euclidean generalizations of algorithms classically defined for data residing in Euclidean spaces. The generalization of machine learning methods to non-Euclidean data first relies on the generalization of their mathematical foundations\textemdash probability and statistics\textemdash  to non-Euclidean spaces. We refer the reader to \cite{Pennec2006} for theoretical foundations on manifolds, and to \cite{Pennec_Sommer_Fletcher_2019} for real-world applications. Beyond probability and statistics, the machine learning algorithms requires non-trivial algorithmic innovations. Such generalizations comprise the bulk of the work reviewed here. 

However, we note that there are two ``simple'' approaches to generalizing Euclidean models that do not \textcolor{black}{require} significant algorithmic innovation. They do not work for all algorithms, and they have limitations. Yet, when possible, they have the benefit of implementational simplicity. We briefly describe two classes of such approaches here\textemdash what we call \textit{``plug-in'' methods} and \textit{tangent space methods}.

\begin{tcolorbox}[colback=lightgray, colframe=black]
\textbf{Non-Euclidean Probability and Statistics.} \\ In a non-Euclidean space, many essential mathematical concepts must be modified to respect the inherent structure of the space. Consider, for example, a dataset consisting of points that lie on the surface of the sphere\textemdash for example, the coordinates of different cities around the globe. The Euclidean mean of the points is a value that lies off-manifold\textemdash a point lying somewhere ``inside'' the sphere but not on the sphere. To find the centroid of these points on the manifold, we must instead use the \textbf{Fréchet mean}, pictured in Fig. \ref{fig:frechet}. This is defined as the point that minimizes the sum of squared \textit{geodesic distances} to all other points in the dataset. By using geodesics, the Fréchet mean is constrained to the manifold, and results in a natural generalization of the notion of a mean to non-Euclidean space. The field of \textbf{Geometric Statistics} defines such non-Euclidean generalizations. We refer the reader to \cite{Pennec2006} for theoretical foundations on manifolds, to \cite{guigui2023introduction} for a comprehensive introduction to these foundations and to \cite{Pennec_Sommer_Fletcher_2019} for real-world applications.
\end{tcolorbox}

\vspace{0.4cm}
\paragraph*{``Plug-In'' Methods} The most straightforward way to generalize machine learning methods to non-Euclidean spaces is to simply replace the definitions of addition, subtraction, distances, and means employed in the Euclidean method with their non-Euclidean counterparts. For example, the k-nearest-neighbors algorithm can be naturally defined for arbitrary non-Euclidean manifolds by replacing Euclidean distance with geodesics. Similarly, k-means can be generalized using geodesic distances and the Fréchet mean. Any \textit{kernel method} that makes use of geodesic distance in the kernel function falls into this category as well. Many popular implementations of machine learning algorithms\textemdash for example in the \texttt{scikit-learn} package\textemdash permit the user specification of a distance function, thus facilitating easy generalization to non-Euclidean spaces. {\textcolor{black}{However, many approaches require deeper modifications that go beyond replacing the operations of the standard method. Spherical CNNs \citep{cohen2018spherical,kondor2018clebsch}, for instance, cannot apply standard convolutions directly, as translations are not well defined on the sphere. Instead, they define convolution by applying group convolutions over the rotation group $SO(3)$ using spherical harmonics, ensuring rotational equivariance and respecting the sphere's geometry.}

}

\vspace{0.4cm}

\paragraph*{Tangent Space Methods} An alternative approach is particularly convenient for non-Euclidean spaces that are manifolds. We call it the tangent space method. It consists in projecting the data from the manifold into the tangent space of a particular point on the manifold, using the so-called \textit{exponential map}. Importantly, the tangent space can be seen as a Euclidean space. Once the data are mapped to Euclidean space, traditional Euclidean machine learning can be applied. This approach typically achieves better results than applying Euclidean methods directly to the original non-Euclidean data. However, in many cases\textemdash particularly for manifolds with greater curvature\textemdash it induces biases in the results due to errors in the local Euclidean approximation of the manifold. Nonetheless, the approach is relatively simple, and can be worthwhile for data lying on manifolds that are flat at the scale of the spread of the data.

Both the plug-in and tangent space methods have the advantage of implementational simplicity. However, many machine learning algorithms require more than just the specification of distances and means, which limits the applicability of the plug-in method. Additionally, in many cases, the biases induced by the tangent space projection are intolerable for the application, which limits its scope. Consequently, in many scenarios, it is necessary to explicitly constrain aspects of the algorithms to the manifolds of interest. This comprises the bulk of the work of Non-Euclidean Machine Learning, which we cover in this section.

We note that the methods reviewed in this paper assume that certain topological, algebraic, or geometric structures are known \textit{a priori} to be present in the data or learning problem. These methods thus require pre-specifying these structures. Sometimes, however, the underlying structure is unknown. A class of methods aims to \textit{discover and characterize unknown non-Euclidean structure} in data. This class includes \textbf{topological data analysis}, certain manifold learning approaches that learn parameters of a latent manifold, such as in \textbf{metric learning}, and algebraic methods for discovering latent group structure, also known as \textbf{group learning}. \textcolor{black}{Other notable approaches include Noether Networks \citep{alet2021noether}, which enforce learned conservation laws, and graph rewiring methods \citep{topping2022understanding}, which dynamically adapt graph connectivity}. These methods are out of scope, as they are used ``prior to'' non-Euclidean machine learning.

\begin{figure*}[hp]
  \centering
  \includegraphics[width=1.0\linewidth]{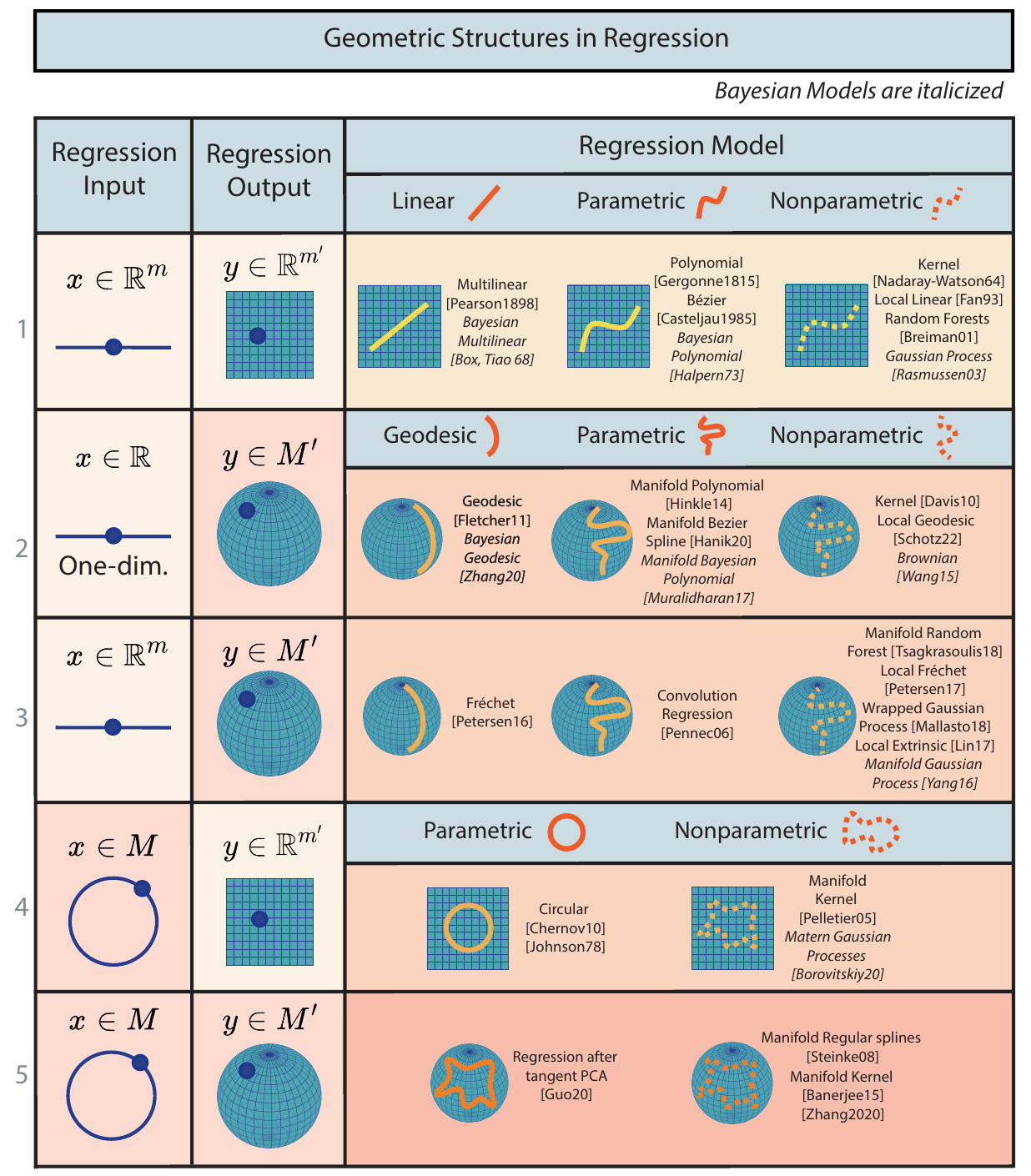}
  \caption{\textbf{Geometric Structures in Regression} categorized according to the geometry of the input and output data spaces (first two columns) and the geometry of the regression model (last column). Yellow boxes correspond to Euclidean space, while orange corresponds to non-Euclidean space. Partially Euclidean cases are light orange. Each pictogram represents the kind of parametrization used: linear (geodesic), nonlinear (nongeodesic) parametric, or nonparametric, with or without Bayesian priors.}
  \label{fig:geo-regression}
\end{figure*}

\subsection{Geometric Structures in Regression}

We first introduce our taxonomy defined by the mathematical structure of both the \textit{data spaces} and the regression \textit{model}. Then, we review the literature on regression. A complete visual representation of the literature is available in Figure~\ref{fig:geo-regression}.

\vspace{0.2cm}

\paragraph*{Taxonomy} In machine learning, regression can be defined as learning a function $f$ going from an input space $X$ to an output space $Y$. Figure~\ref{fig:geo-regression} organizes regression models into a taxonomy based on the geometric properties of input and output spaces\textemdash see first two columns in Figure~\ref{fig:geo-regression}. Here, we distinguish conventional Euclidean spaces from more complex manifold spaces, setting the stage for a detailed exploration of five key configurations: Euclidean to Euclidean, one-dimensional Euclidean to Manifold, Euclidean to Manifold, Manifold to Euclidean, and Manifold to Manifold. 

Each configuration is then distinguished by its regression model\textemdash see third column in Figure~\ref{fig:geo-regression}. \textit{Linear} methods assume a relationship between the input variable $x \in X$ and output variable $y \in Y$ that can be represented by a linear function as \( y = \mathbf{A}x + \mathbf{b} \), where \( \mathbf{A} \) is a matrix of coefficients, and \( \mathbf{b} \) is a vector of constants\textemdash hence constraining the $y$s to belong to a linear space characterized by the parameters $A, b$. We note that linear regressions are not appropriate on manifolds since the addition operation, a linear operation, is not well-deﬁned on a manifold, a non-linear space. Consequently, linear methods become \textit{geodesic} methods on manifolds on rows 2 and 3 of Figure~\ref{fig:geo-regression}, where the relationship between $x$ and $y$ can be represented by a geodesic characterized by a set of parameters analogous to $A, b$ above. Next, \textit{non-linear parametric} and \textit{non-geodesic parametric} methods involve relationships described by a fixed set of parameters but in a more complex form, such as polynomials or other non-linear equations $y = f_\theta(x)$ where $\theta$ represents the parameters. \textit{Non-parametric} approaches, on the other hand, do not assume a parametric form for the relationship between $x$ and $y$, providing flexibility to model relationships that are directly derived from data. The term nonparametric does not necessarily mean that such models completely lack parameters but that the number and nature of the parameters are flexible and not fixed in advance, contrarily to parametric approaches. Further, these methods are \textcolor{black}{distinguished} as \textit{Bayesian} or \textcolor{black}{\textit{frequentist}\textemdash as denoted by the italicized models} in Figure~\ref{fig:geo-regression}. \textit{Bayesian} methods integrate prior probabilistic distributions with observed data, facilitating the updating of knowledge about the parameters in light of new evidence. This approach contrasts with \textcolor{black} {the remaining methods, which rely exclusively on observed data, typically employing so-called frequentist statistical principles without incorporating prior distributions on parameters.}

In what follows, we review regression models according to this geometric taxonomy, row by row in Figure~\ref{fig:geo-regression}. For each category of regression models, we showcase one emblematic, category-defining, paper that reflects the transition from traditional Euclidean analysis to the manifold paradigm. We consider the following papers to be \textit{out of scope}: 1) Papers that generalize a class of curves (e.g., splines) on manifolds, but do not leverage this generalization to perform regression, 2) Papers that perform interpolation on manifolds, but not regression, 3) Papers whose regression method has been developed for only one type of manifold (e.g., only for spheres).

\begin{enumerate}
    \item \textbf{Euclidean Input, Euclidean Output:} We review classical regression models that have been generalized to manifolds, in the first row of Figure~\ref{fig:geo-regression}. Initiated with linear regression by Legendre and Gauss \citep{legendre1805}, this category has expanded to include nonlinear parametric models, particularly polynomial models introduced by Gergonne \citep{gergonne1815}. The inclusion of non-parametric methods is marked by the Nadaraya-Watson kernel methods \citep{nadaraya1964estimating}, further local linear models \citep{fan1993}, and Breiman's development of random forests \citep{breiman2001}. In Bayesian analysis, the linear and polynomial approaches are respectively represented by Bayesian Multilinear \citep{box1968} and Polynomial Bayesian models \citep{halpern1973}, with the Gaussian Process \citep{RasmussenGaussianLearning} illustrating the non-parametric Bayesian perspective.

    \item \textbf{One-dimensional Euclidean Input, \textcolor{geometry}{Manifold Output:}} The earlier generalizations of classical regression models to manifolds involve generalizing the output space. Many of these models consider a one-dimensional input $x$, and are shown in the second row of Figure~\ref{fig:geo-regression}. Geodesic regression, \textcolor{black}{a manifold generalization of linear models introduced nearly 200 years later,} handles one-dimensional Euclidean inputs \citep{fletcher2011geodesic}. Polynomial regressions on manifolds \citep{hinkle2012} and Bezier-splines fitting on manifolds \citep{Hanik2020NonlinearRO} generalize their Euclidean counterparts to manifolds in the output space, 197 years and 35 years later respectively. Non-parametric methods with output values on manifolds include Fréchet-casted Nadaraya-Watson kernel regression \citep{davis2010}, and local geodesic regression \citep{schotz2022}\textemdash the latter being the counterpart of local linear regression to manifolds. In terms of Bayesian methods, the geodesic regression model was made Bayesian in \citep{zhang2020}, while manifold polynomial regression turned Bayesian in \cite{muralidharan2017polynomial}. Lastly, one non-geodesic, non-parametric, Bayesian model belongs to this category: kernel regressions by Devito and Wang's Brownian motion model \citep{wang2015} which generalizes Nadaraya-Watson's approach 51 years later.

    \item \textbf{Euclidean Input, \textcolor{geometry}{Manifold Output:}} Next, we review regression models with outputs on a manifold, for which the input is not restricted to be one-dimensional, in the third row of Figure~\ref{fig:geo-regression}. 
    The Fréchet regression \citep{petersen2016frechet} generalizes the geodesic regression to higher-dimensional Euclidean inputs. We also find several nongeodesic parametric regression models. One of the earliest works in this category deals with the parametric regression of regularized manifold valued functions \citep{Pennec2006b}. A key idea is to rephrase convolutions as weighted Fréchet means of manifold variables, which become the parameters of the implicit function. 
    Also in the nongeodesic parametric regression category, we find the semi-parametric intrinsic regression model by \citep{Shi2009}, or the stochastic development regression by \cite{kuhnel2017}. Non-parametric methods with manifold-valued outputs include the manifold random forests \citep{tsagkrasoulis2018} that generalize their Euclidean counterpart 15 years later, the local Fréchet regression~\citep{petersen2016frechet}, another multi-dimensional generalization of the local linear regression, and the local extrinsic regression by \citep{lin2017}. In terms of Bayesian approaches, we observe a lack of methods within geodesic and nongeodesic parametric models. However, we find Bayesian non-parametric methods, such as Manifold Gaussian processes  \citep{yang2016bayesian} and Mallasto's wrapped Gaussian processes \citep{mallasto2018wrapped} which generalizes Williams and Rasmussen's Gaussian processes 9 and 13 years later.

    \item \textbf{\textcolor{geometry}{Manifold Input}, Euclidean Output:} We now review geometric generalizations of Euclidean regression models that have received less attention: models in which the regression input space is a manifold. We start with methods for which the output space is a Euclidean space, in the fourth row of Figure~\ref{fig:geo-regression}. Johnson and Wehrly defined a regression model for an angular (one-dimensional) variable with a Euclidean scalar (one-dimensional) output variable \citep{johnson1978}.
    Chernov's circular regression outputs to higher-dimensional Euclidean spaces, typically two- or three-dimensional \citep{chernov2010}. Pelletier's non-parametric regression approach stands out for estimating functions from manifold inputs to Euclidean outputs \citep{Pelletier2006NonparametricRE}. Both methods are non-Bayesian. In terms of Bayesian methods, we find the Bayesian circular-linear regression method \citep{gill2010}, \textcolor{black}{and the Bayesian non-parametric method Matérn Gaussian Processes on manifolds \citep{NEURIPS2020_92bf5e62}}.

    \item \textbf{\textcolor{geometry}{Manifold Input, Manifold Output:}} The fifth row of Figure~\ref{fig:geo-regression} presents the most general case of regression, from a geometric perspective. In these models, both the regression input and output spaces are manifolds. This category includes a method that first transforms both input and output data from manifolds to Euclidean spaces, and then apply an auxiliary classical regression model on Euclidean spaces \citep{Guo2019}. In the nonparametric methods, we find Steinke's regular splines methods~\citep{steinke2008} and Banerjee's manifold kernel regression~\citep{banerjee2015} generalizing in 2015 both the classical kernel regression from 1964 and its generalization to manifold output space from 2010. At the time of the review, there is no method in this category that adopts the Bayesian point of view.

\end{enumerate}

\begin{figure*}
  \centering
  \includegraphics[width=1.0\linewidth]{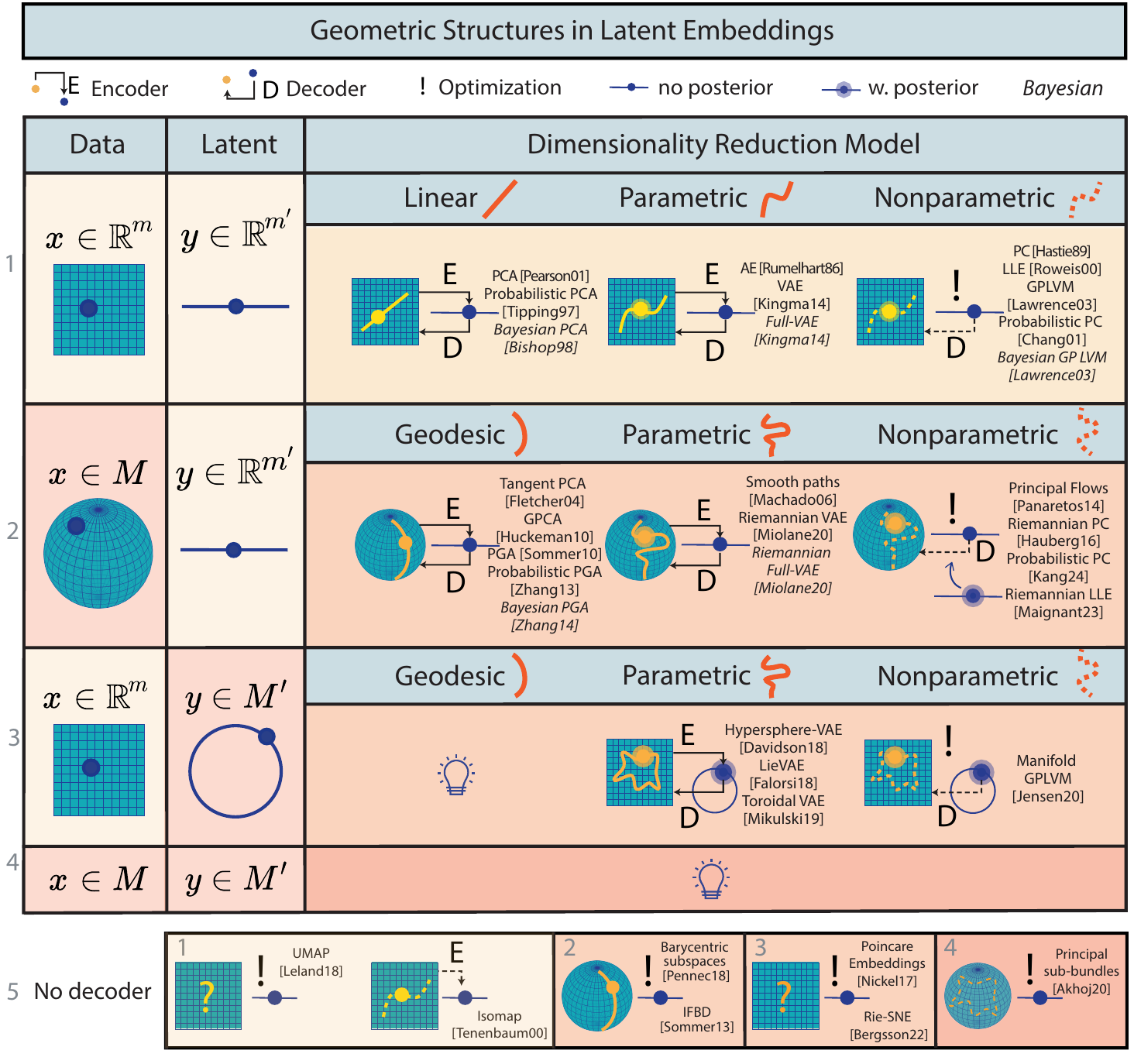}
  \vspace{0.3cm}
  \caption{\textbf{Geometric Structures in Latent Embeddings} categorized according to the geometry of the data and latent spaces (first two columns) and of the latent embedding model (last column). Yellow boxes correspond to Euclidean space, while orange corresponds to non-Euclidean space. Partially Euclidean cases are light orange. Each model is further classified by use and type of encoder/decoder, as well as whether it computes a posterior on the latents, and whether it is Bayesian. (P)PC: (Probabilistic) Principal Curves; GP LVM: Gaussian Process Latent Variable Model; PGA: Principal Geodesic Analysis; GPCA: Geodesic PCA. \vspace{-0.5cm}
  \label{fig:geo-dimension-reduction}}
\end{figure*}

\subsection{Geometric Structures in Latent Embeddings}

We first introduce our taxonomy
defined by the mathematical structure of both the \textit{spaces} and the latent embedding \textit{model}. Then, we review the literature on latent embeddings in Figure~\ref{fig:geo-dimension-reduction}, with details in the text.

\vspace{0.4cm}
\paragraph*{Taxonomy} We define the problem of latent embeddings as transforming (typically, high-dimensional) data from their data space $X$ into a latent space $Y$ (typically, of lower dimension, i.e., $\dim(Y) < \dim(X)$). Figure~\ref{fig:geo-dimension-reduction} organizes latent embeddings methods into a taxonomy first based on the geometric properties of the data and latent spaces\textemdash see the first two columns. It differentiates between conventional Euclidean spaces and the more complex manifold spaces, setting the stage for the following four key configurations: Euclidean Data to Euclidean Latents, Manifold Data to Euclidean Latents, Euclidean Data to Manifold Latents, and Manifold Data to Manifold Latents. For each configuration, the (usually, lower dimensional) latent space $Y$ is schematically represented as a space of dimension 1: a line for Euclidean spaces, and a circle for manifolds. The (usually, higher-dimensional) data space $X$ is schematically represented as a space of dimension 2: a plane for Euclidean spaces, and a sphere for manifolds. Yet, we emphasize that we review all latent embedding methods here; not only the methods going from dimension 2 to dimension 1.

Each configuration is then distinguished by the approach to latent embedding\textemdash see third column in Figure~\ref{fig:geo-dimension-reduction}. First, latent embedding approaches are organized depending on whether they leverage a decoder, and if so, of what type: linear (resp. geodesic), nonlinear (resp. nongeodesic) parametric, non-parametric\textemdash represented by full line, full curved line, and dashed curved line in the pictograms of Figure~\ref{fig:geo-dimension-reduction}. While every method converts data into latents, only some methods introduce a decoder $D$ that converts latents $y$s back into data $x$'s: $x = D(y)$. When a decoder $D$ exists, the latent space $Y$ can be mapped into the data space $X$ via $D(Y) \subset X$. The yellow and orange curves in the pictograms represent that mapping. We further distinguish two subcases: whether the mapped space $D(Y)$ is linear (geodesic, in the manifold case), or nonlinear (or non-geodesic). The presence of a decoder is represented by a black arrow and the legend $D$; the black arrow is a dotted arrow if the decoder is non-parametric. Methods that do not leverage any decoder are reviewed independently at the bottom of Figure~\ref{fig:geo-dimension-reduction}.

Second, approaches are organized depending on whether they leverage an encoder $E$, where $E$ is a function mapping data $x$'s to latents $y$s: $E(x) = y$. Indeed, while every method converts data into latents, only some methods do so through an explicit encoder function $E$. Others might only compute the latent $y_i$ corresponding to a data point $x_i$ through the result of an optimization $\text{min}_{y} \text{Cost}(x_i)$. The presence of an encoder is represented by a black arrow and the legend $E$, and the arrow is a dotted arrow if the encoder is non-parametric. The use of optimization is represented by the $!$ and no black arrow.

Third, approaches are categorized based on how they compute uncertainty on the decoder parameters, i.e., whether they are Bayesian or not. For example, traditional PCA does not incorporate uncertainty, but Bayesian PCA does. \textcolor{black}{Bayesian approaches are distinguished by text in \textcolor{black}{italics}.}

We further explain if the data is assumed to come from a generative model. A generative model explains how data points are generated from latents using probability distributions. For example, principal component analysis (PCA) has a decoder, but no generative model; while probabilistic PCA uses a decoder with a generative model. We also note if approaches compute an uncertainty on the latents $y$ associated with the data points $x$. Uncertainty on the latents is represented by the posterior distribution $p(y|x)$.

We now survey the various categories of latent embeddings row by row in Figure~\ref{fig:geo-dimension-reduction}.

\begin{enumerate}

\item \textbf{Euclidean Data, Euclidean Latents:} We describe approaches from this configuration subcolumn by subcolumn. Principal Component Analysis (PCA) \citep{pearson1901liii} learns a linear subspace, while Probabilistic PCA (PPCA) \citep{tipping1999probabilistic} achieves the same goal within a probabilistic framework relying on a latent variable generative model, which provides a posterior distribution on the latents. Bayesian PCA \citep{bishop1998bayesianp} additionally learns a probability distribution on the parameters of the models, representing the parameters of the linear subspace (e.g., slope and intercept). These methods are restricted in the type of subspace that can be fitted to the data: only linear subspaces. To lift this restriction, we also find numerous methods that learn nonlinear manifolds from Euclidean data. The whole field of manifold learning fits in this case, and can be subdivided into further subcategories depending on \textit{how} the learned manifold is being represented: by a nonlinear parametric decoder, by a nonparametric decoder, or without any decoder at all. Here, we only introduce one category-defining method for each of the subcategories we consider.

In the nonlinear parametric decoder category, we introduce autoencoders. Autoencoders (AEs) \citep{rumelhart1986learningrb} learn a nonlinear subspace of a Euclidean space, while variational autoencoders (VAEs) \citep{Kingma2014Auto-EncodingBayes} achieve the same goal with a probabilistic framework relying on a latent variable generative model. The Full-VAE model proposed in \citep{Kingma2014Auto-EncodingBayes} additionally learns a posterior on the parameters of the model, i.e., the parameters of the decoder. The methods in this category all leverage an encoder. In the nonparametric decoder category, we introduce the generative model principal curves \citep{Hastie1989PrincipalCurves}, which fit a nonlinear manifold to the data, but do not leverage a posterior on the latents. In this same category, we also introduce Local Linear Embedding (LLE) \citep{Roweis2000NonlinearEmbedding} and Gaussian Process Latent Variable Models (GP LVM) \citep{lawrence2003gplvm}, which all provide posteriors on the latents. A probabilistic approach to principal curves (PPS) developed in \cite{Chang2001ASurfaces} also falls under this category. Finally, we introduce the Bayesian GP LVM, which is generative and additionally provides a posterior on the model's parameters. We note that the methods of this category do not leverage any encoder, and instead compute the latent associated with a given data point by solving an optimization problem.

These techniques are based on vector space operations that make them unsuitable for data on manifolds. Consequently, researchers have developed methods for manifold data, which take into account the geometric structure; see next row in the Table, described in the next paragraph. 

\item \textbf{\textcolor{geometry}{Manifold Data}, Euclidean Latents:} We describe approaches subcolumn by subcolumn. In the geodesic decoder category, Principal Geodesic Analysis (PGA) \citep{fletcher2004principal, sommer2014optimization}, tangent PGA (tPGA) \citep{fletcher2004principal}, and Geodesic Principal Component Analysis (GPCA) \citep{Huckemann2010IntrinsicActions}
learn variants of geodesic subspaces, generalizing the concept of a linear subspace to manifolds. As such, these methods represent different generalizations of PCA to manifolds. Probabilistic PGA \citep{Zhang2013ProbabilisticAnalysis} achieves the same goal, while adding a latent variable model generating data on a manifold, and hence generalizes probabilistic PCA, 16 years later. Similarly, Bayesian PGA \citep{Zhang2013ProbabilisticAnalysis} generalizes Bayesian PCA by including the posterior distribution of the parameters defining the submanifolds, i.e., the base point and tangent vectors defining the principal (geodesic) components. 
However, these methods are restricted in the type of submanifold that can be fitted to the
data, that is: geodesic subspaces at a point.

The restriction to globally defined subspaces based on geodesics can be considered
both a strength and a weakness. While it protects from the problem of overfitting with a submanifold that is too flexible, it also prevents the method from capturing possibly nonlinear effects. With current dataset sizes exploding (even within biomedical imaging datasets which have been historically much smaller), the investigation of flexible submanifold learning techniques may become increasingly important.

In the nongeodesic parametric decoder category, we consider various generative models. A natural extension with one-dimensional latents is to define splines using higher order polynomials that are then fitted to a set of points on a Riemannian manifold \citep{machado_fitting_2006,machado_higher-order_2010} or on a Lie group \citep{gay-balmaz_invariant_2012}. We then consider higher dimensional Euclidean latents. Variational autoencoders have been generalized to manifold data in \citep{miolane2020learning}, a methodology that can be applied to both AEs, VAEs and Full-VAEs on manifolds. The AEs, VAEs and Full-VAEs are the only methods that learn   a multidimensional nongeodesic submanifold parameterized with a latent variable model. 

In the nongeodesic, nonparametric decoder category with one-dimensional Euclidean latents, principal flows \citep{Panaretos2014PrincipalFlows} and Riemannian principal curves \citep{Hauberg2016PrincipalManifolds} generalize traditional Euclidean principal curves to Riemannian manifolds, 25 years later. The probabilistic Riemannian principal curves \citep{kang_probabilistic_2024} further introduce a probabilistic framework relying on a latent variable generative model, hence generalizing the Euclidean probabilistic principal curves, 23 years later. Lastly, for higher dimensional Euclidean latents: the Riemannian LLE \citep{Maignant2023}, a generative model which generalizes its Euclidean counterpart, the LLE, 23 years later. We note the absence of works performing latent embedding via a nonparametric decoder in a generative model nor in a Bayesian framework. This represents a possible avenue for research.

\item \textbf{Euclidean Data, \textcolor{geometry}{Manifold Latents.}} We describe approaches subcolumn by subcolumn. We first find approaches that belong to the VAE framework, where the latent space is a manifold, even though the data belong to a Euclidean space. For example, the hypersphere VAE by \citet{Davidson2018HypersphericalAuto-encoders} proposes a hyperspherical latent space, \citet{falorsi2018homeomorphic} propose a Lie group latent space, and \citet{mikulski2019toroidal} propose a toroidal latent space. All of these are generative approaches which provide an (approximate, amortized) posterior on the latent variables, represented as a probability distribution on the manifold of interest. In some sense, these approaches represent the counterpart of \citet{miolane2020learning} which considers a manifold data space and a Euclidean latent space. Next, we find nonparametric decoder approaches, such as the manifold Gaussian Process Latent Variable Model (GPLVM) \citep{jensen2020} which generalizes the GPLVM from the Euclidean case \citep{lawrence2003gplvm}.

\item \textbf{No Decoder.} Few models avoid using a decoder. For this reason, we briefly survey these models across all geometric data and latent structures. In the Euclidean data with Euclidean latent case (corresponding to row 1), we find Uniform Manifold Approximation and Projection (UMAP) \citep{McInnes2018} and Isomap \citep{Tenenbaum2000}. These approaches learn lower-dimensional representations of data but do not provide a latent variable generative model, nor a parameterization of the recovered subspace. In the manifold data with Euclidean latent case (corresponding to row 3), a flexible generalization of linear subspaces to manifolds is given by barycentric subspaces \citep{Pennec2018BarycentricManifolds}, where the submanifold is defined implicitly through geodesics to several reference points. 
Iterated Frame Bundle Development (IFBD) \citep{sommer_horizontal_2013} is another optimization method that iteratively builds principal coordinates along new directions. \textcolor{black}{GEOMANCER \citep{pfau2020disentangling} also falls into this category: it discovers disentangled latent factors by analyzing how local tangent spaces transform under parallel transport. This non-parametric method constructs subspace-valued features via spectral diffusion geometry, revealing latent product structure.} In the Euclidean data with manifold latent case (row 4), we first embed data points in $X$ into a hyperbolic latent space $Y$ using Poincaré embeddings \cite{nickel2017poincare}. Hyperbolic space is useful for representing data with hierarchical structure. 
Second, the Riemannian SNE \citep{bergsson2022visualizing} generalizes traditional Euclidean SNE \citep{vanDerMaaten2008} but 14 years later. Finally, the last decoder-free model, called principal subbundles \citep{akhoj2023}, is the only model designed for manifold data and manifold latents. Indeed, we note that there are no decoder-based approaches for latent embeddings for this case. 
Overall, the sparsity of decoder-free latent embedding models leaves open opportunities for many choices of geometric data/latents.

\end{enumerate}

\subsection{\textcolor{black}{Topological Structures in Regression}}

\begin{figure*}[t]
    \centering
    \includegraphics[width=1.0\linewidth]{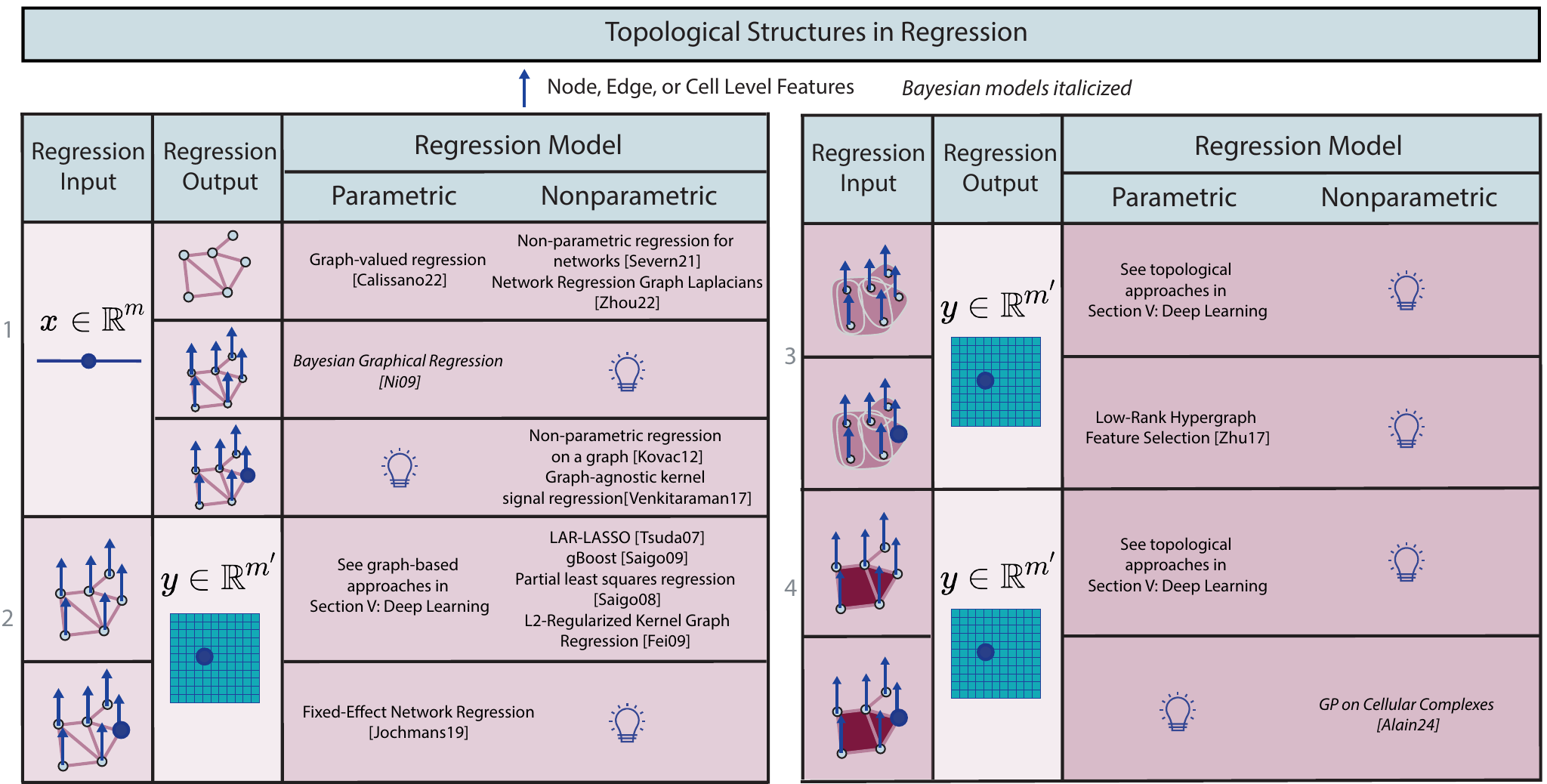}
    \caption{\textbf{Topological Structures in Regression} categorized according to the topology of the input and output data spaces (first two columns) and the kind of parametrization of the model (last two columns).}
    \label{fig:topo-regression}
\end{figure*}

\textit{Taxonomy:} To complement the \emph{geometric} view of Figure~\ref{fig:geo-regression}, Figure~\ref{fig:topo-regression} classifies regression models whose input space $X$, output space $Y$, or both are \emph{topological structures}. This is specified in the first two columns of both tables. Here, we distinguish conventional Euclidean spaces from more complex topological spaces, setting the stage for a detailed exploration of four configurations: Euclidean to Graph, Graph to Euclidean, Hypergraph to Euclidean, Cell Complex to Euclidean.

For each configuration the figure further distinguishes \emph{labeled} complexes, where node-, edge- or cell-level features are provided, from \emph{unlabeled} ones that encode only incidence information, by dividing each input/output pair into two rows. When features are present we specify in the text the exact level(s) on which they reside.

The configurations are similarly distinguished by their regression models and separated by whether they are \textit{parametric} or \textit{non-parametric}, see third column of both tables. Unlike Figure~\ref{fig:geo-regression}, the topology table omits a \textit{Linear/Geodesic} column because concepts such as straight lines or geodesics require a metric structure that generic topological objects (e.g., graphs, hypergraphs, or complexes) do not possess.

Further, these methods are distinguished as \textit{Bayesian} or \textit{frequentist}\textemdash as denoted by the italicized works in Figure~\ref{fig:topo-regression}. 

It is important to note that depending on the task, the regressor may output the \emph{entire} topological object, such as a full graph, hypergraph, or complex, or more granular elements like node- or edge-level predictions. The accompanying text specifies the level of the output of the regression models.

In what follows, we review regression models according to this topological taxonomy, row by row in Figure~\ref{fig:topo-regression}. We consider the following papers to be \textit{out of scope}: Papers that perform interpolation or classification on topological structures, but not regression.

\begin{enumerate}
    \item \textbf{Euclidean Input, \textcolor{topology}{Graph Output}:} A first class of methods predicts the structure of entire graphs, either using parametric models \citep{Calissano22} or non-parametric alternatives \citep{Severn21,Zhou22}. When the predicted graph includes node-level features, Bayesian graphical regression offers a probabilistic framework \citep{ni2009}. Shifting focus from entire graphs to individual nodes, a distinct line of work assumes the graph structure is known and uses it as a regularizer to predict node-level features \citep{kovac12,venkitaraman17}.

    \item \textbf{\textcolor{topology}{Graph Input}, Euclidean Output:} When predicting from graphs with features, parametric methods such as graph neural networks arise -- see Section~\ref{sec:dl}. Non-parametric approaches decompose the graph into subgraphs and apply various regression models \citep{Tsuda07,Saigo08a,Saigo08b,Fei09}. We note that, even for unfeatured graphs, one can define basic node features—such as constant values, node degrees, or structural properties—to enable feature-based prediction. Last, a distinct line of work predicts at the node level, i.e, the graph is known and is used as a regularizer of the regression in \citep{Jochmans19}.

    \item[\hspace{-1.25em}3)] \textbf{\textcolor{topology}{Hypergraph Input}, Euclidean Output:} As with graphs, the regression can predict from featured hypergraphs. Parametric models, including hypergraph neural networks, address this task (see Section~\ref{sec:dl}), while non-parametric methods remain unexplored, offering a promising direction for future work. Even without features, structural properties of the hypergraph can be used to define them. Finally, a distinct class of methods uses a single node feature as input of the regression, using the hypergraph as a regularizer \citep{Zhu17}.

        \item[\hspace{-1.25em}4)] \textbf{\textcolor{topology}{Cell Complex Input}, Euclidean Output:} The regression can also predict from featured simplicial and cellular complexes, with parametric models including simplicial and cellular neural networks (see Section~\ref{sec:dl}) and non-parametric methods unexplored. A method using a single node feature as input of the regression is presented in a Bayesian framework in \citep{alain23}.

\end{enumerate}

\subsection{Topological Structures in Latent Embeddings}

\begin{figure*}[t]
    \centering
    \includegraphics[width=1.0\linewidth]{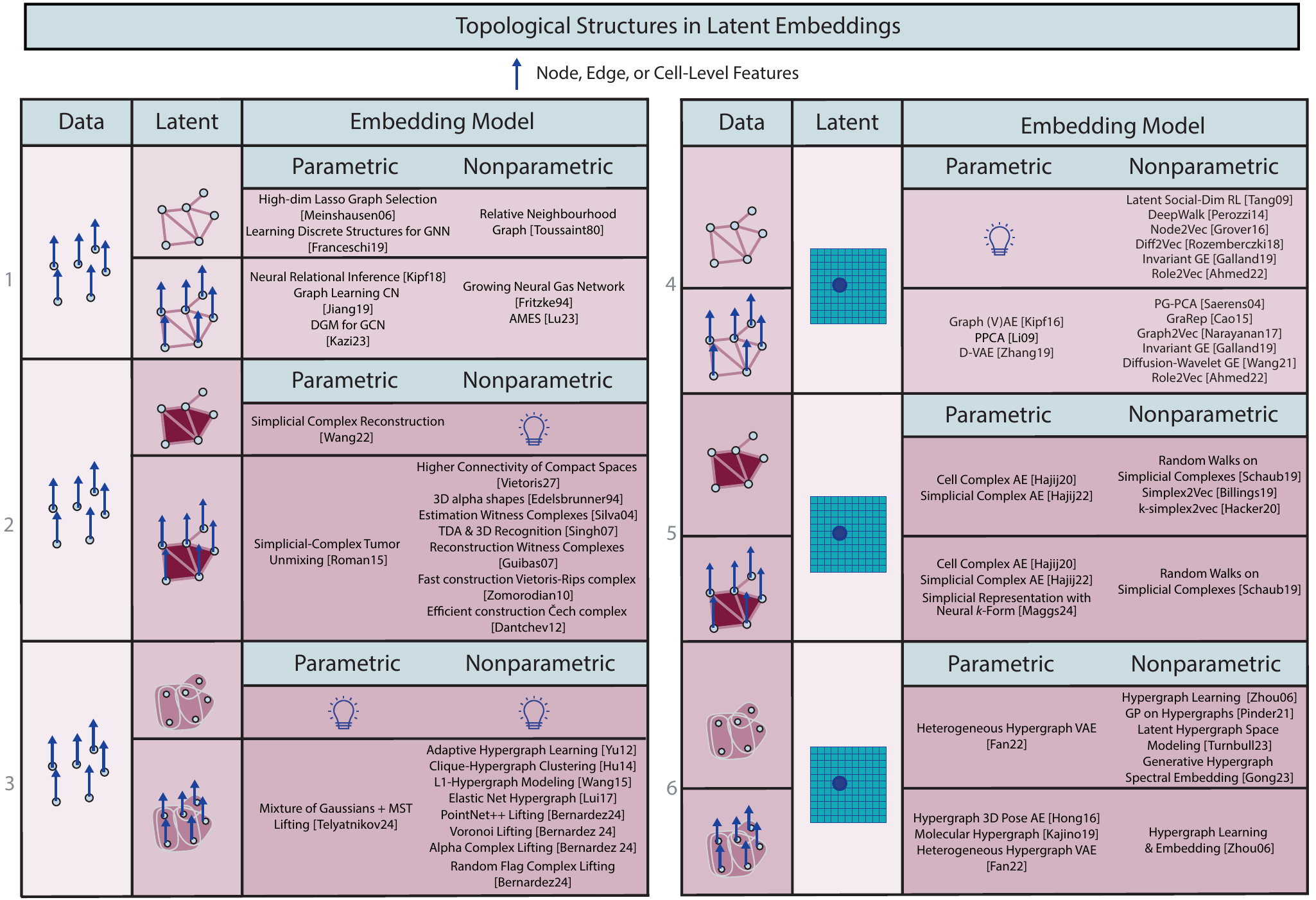}
    \caption{\textbf{Topological Structures in Latent Embeddings} categorized according to the topology of the input and output data spaces (first two columns), the kind of parametrization of the model (last two columns), and the nature of labels: complex-level or node-level (two rows per box).}
    \label{fig:topo-latent-embedding}
\end{figure*}

\textit{Taxonomy:} In this setting, we note that the embedding approach relocates data between Euclidean and topological domains without necessarily lowering dimension-- changing the representation rather than necessarily compressing it. Figure~\ref{fig:topo-latent-embedding} therefore classifies latent embedding methods first by the nature of the data and latent spaces, see the first two columns, covering six source–target configurations: Point Cloud to Graph, Point Cloud to Simplicial or Cellular Complex, Point Cloud to Hypergraph, Graph to Euclidean, Simplicial or Cellular Complex to Euclidean, and Hypergraph to Euclidean. We do not include Euclidean to Euclidean in this table because that is covered in Figure \ref{fig:geo-dimension-reduction}. The review in \citep{hensel2021survey} offers additional details.

For each configuration the figure further distinguishes \emph{labeled} complexes, where node-, edge- or cell-level features are provided, from \emph{unlabeled} ones that encode only incidence information, by dividing each input/output pair into two rows. When features are present, we specify in the text the exact levels (node, edges, etc) on which they reside. Models that can operate with or without features are included in both rows. Each configuration is then distinguished by the approach to latent embedding\textemdash see third column in Figure~\ref{fig:topo-latent-embedding}, depending on whether they are parametric or non-parametric. We do not include a \textit{Linear} column for the same reasons as in the previous section.

We once again note that depending on the task, the latent embedding model may input or embed the \emph{entire} topological object, such as a full graph, hypergraph, or complex, or more granular elements like nodes, edges or cells. The accompanying text specifies the level of inputs or embeddings for each model.

We now survey the various categories of latent embeddings row by row in Figure \ref{fig:topo-latent-embedding}. We consider the following ``lifting'' methods to be \textit{out of scope}: methods that go from a graph to a topological domain, or between topological domains, and refer to \citep{topobenchmarkx2024} for their review.

\begin{enumerate}
    \item \textbf{\textcolor{topology}{Point Cloud Input, Graph Output}:} We describe approaches from this configuration starting with unlabeled graphs and then moving onto labeled. Within the unlabeled graphs, the majority of latent embedding approaches are node level, meaning that each point in the point cloud becomes a node in the outputted graph. Node level parametric approaches include graph construction with Lasso regression \citep{10.1214/009053606000000281} bilevel Bernoulli-edge learning (LDS) and \citep{franceschi2019learning}. For nonparametric unlabeled approaches, we have the node level approach \cite{TOUSSAINT1980261}.

     For labeled outputs, parametric node level approaches include the neural relational inference model \cite{kipf2018neural} that utilizes a variational GNN and learns features on edges, \cite{8953909} which jointly learns an adaptive adjacency via a trainable similarity kernel inside a graph-learning–convolutional network and features on nodes, and \cite{9763421} which inserts a Gumbel-Softmax edge-sampling module to learn task-specific graph structure end-to-end and features on nodes. Nonparametric node level labeled approaches include the growing neural gas network which outputs a graph with node and edge features \citep{NIPS1994_d56b9fc4} and Attentional Multi-Embedding Selection \citep{lu2023ames}.

    \item \textbf{\textcolor{topology}{Point Cloud Input, Simplicial/Cellular Complex Output:}} For unlabeled parametric models, reconstruction of simplicial complexes from binary contagion and Ising data constructs a simplicial complex where points become the 0-simplices of the complex \citep{wang2022full}. No unlabeled nonparametric methods exist, indicating an opportunity for further exploration.

    In terms of labeled parametric models, \cite{roman2015simplicial} is a component level approach that clusters the point cloud, then fits a simplex to each cluster and treats the simplex vertices as the genomic profiles of un-observed cell types. For non parametric models, node-level the Vietoris–Rips complex \citep{vietoris1927hoeherer} outputs a simplicial complex with original features on nodes and no features on edges nor simplices. An efficient approach to this construction is presented in \citep{ZOMORODIAN2010263}. \cite{10.1145/174462.156635} outputs a sequence of simplicial complices, with the original features on the nodes. \cite{DANTCHEV2012708} does the same but outputs a single simplicial complex. Similarly, \cite{:10.2312/SPBG/SPBG04/157-166} and \cite{10.5555/1283383.1283499} have the same structure for a single complex but with only a subset of the original features. Original points are clustered and then those clusters become a 0-simplex in the complex in \citep{:10.2312/SPBG/SPBG07/091-100}.
    
    This selection is not exhaustive since all of topological data analysis and persistence homology would fall here.

    \item \textbf{\textcolor{topology}{Point Cloud Input, Hypergraph Output:}} Beginning with unlabeled outputs, no methods exist—indicating an
    opportunity for further exploration. For labeled outputs, a parametric method is the Mixture of Gaussians MST lifting,  transforming a point cloud into a hypergraph by employing a Gaussian mixture model and constructing a minimal spanning tree between the generated Gaussians \cite{topobenchmarkx2024}.  Nonparametric methods include \cite{6165360}, \cite{hu2014high}, \cite{7064739} and \cite{10.1109/TIP.2016.2621671}, which are node-level and every inputted point becomes a hypergraph vertex, and hyperedges receive learned weights. The remaining nonparametric methods include the feature-based Vonoroi, PointNet++, Alpha Complex and Random Flag Complex liftings implemented in \cite{topobenchmarkx2024}.

    \item \textbf{\textcolor{topology}{Graph Input}, Euclidean Output:}
    No known parametric unlabeled approaches exist. \texttt{Node2vec} \citep{10.1145/2939672.2939754}, \texttt{Deepwalk} \citep{10.1145/2623330.2623732}, \texttt{Diff2Vec} \citep{rozemberczki2018fast} and \cite{10.1145/1557019.1557109} perform node-level graph embeddings. Graph-level embedding models include invariant embedding \citep{galland:hal-02947290}. \texttt{Role2Vec} embeds the vertex-types (i.e., structural “roles”) that multiple vertices are first mapped into \citep{9132694}.
    
    Parametric labeled graph models include the Graph Autoencoder and graph Variational autoencoder \cite{kipf2016vgae}, which can be made Bayesian by using the Full Graph Variational autoencoder. These models implement graph node and feature embedding. {D-VAE} introduces an graph embedding autoencoder for directed acyclic graphs with node features \citep{10.5555/3454287.3454429}. Probabilistic Relational PCA performs graph node embedding using a probabilistic framework \citep{NIPS2009_69adc1e1}. In terms of nonparametric models, \texttt{Role2Vec}, \texttt{Graph2Vec} and invariant embedding can accomodate graphs with or without features. Principal component analysis of a graph extends PCA to weighted graph structures, embedding graph nodes \citep{saerens2004pca-graph}. \texttt{GraRep} embeds graph nodes for graph with edge weights and no node features \citep{10.1145/2806416.2806512}. For full graph embedding, \citep{10.1145/3459637.3482115} implement a diffusion-wavelet characterization of node-feature distributions for graphs with node features only.


    \item \textbf{\textcolor{topology}{Simplicial/Cellular Complex Input}, Euclidean Output:} Parametric methods for embedding entire simplicial or cell complexes, with or without features, have been developed through simplicial and cellular autoencoders \citep{hajij20,hajij22}. When features are absent, structural attributes of the complex can be used to define node features. Among non-parametric approaches, extensions of \texttt{node2vec} to the simplicial domain have been proposed, including \texttt{simplex2vec} \citep{billings19} and \texttt{ksimplex2vec} \citep{hacker20}, both relying on random walks, an idea also explored in \citep{schaub19}. \textcolor{black}{Neural \(k\)-Forms \citep{maggs2024simplicial} represent simplicial complexes as Euclidean embeddings by integrating learnable differential forms over the embedded \(k\)-simplices. Thus, this method builds vector representations through geometric integration.}

    \item \textbf{\textcolor{topology}{Hypergraph Input}, Euclidean Output:} Among methods embedding hypergraphs with or without features, we find the heterogeneous hypergraph VAE (parametric) \citep{fan22}, with a nonparametric alternative in \citep{zhou06}. Other parametric approaches such as \citep{hong16,kajino19} require features on the hypergraphs' nodes, while other nonparametric approaches exclusively embed hypergraph structures \citep{pinder21,turnbull21,gong23}.

\end{enumerate}

This concludes our review and categorization of non-Euclidean machine learning methods in regression and latent embeddings. For a more detailed review of topology in machine learning, we refer the reader to \citep{hensel2021survey}. While several deep neural network models\textemdash such as variational autoencoders\textemdash appear in our latent embedding taxonomy, we have saved a more complete treatment of deep learning for the next section. Deep neural networks stand out from more traditional machine learning methods in that they are flexible compositions of functions that progressively transform data between spaces. In our consideration of latent embedding methods, we abstracted away from the transformations performed by individual neural network layers to consider only the structure of the input and latent spaces (which may be many layers deep in a model). In the next section, we explicitly consider the input-output structure of individual neural network layers, and the ways in which topology, geometry, and algebra have been incorporated into these layers.

\section{Survey: Non-Euclidean Deep Learning}\label{sec:dl}

We now review non-Euclidean structures in deep learning, particularly focusing on how geometry, topology, and algebra can enrich the structure of a given layer within a deep neural network. A neural network layer is a function $f: X \rightarrow Y$, and thus can be analyzed and categorized in terms of the mathematical structure of the input space $X$ and output space $Y$. We first cover neural network layers without an attention mechanism ({\color{black}see the first paragraph in Section \ref{sec:attention_mech} for an introductory description of attention}), followed by layers with attention.  \textcolor{black}{We note that we will focus on the explicit mathematical structure of the input and output spaces of each neural network layer, rather than the emergent properties of their latent representations --leaving the analysis of the latter to future work.}

\subsection{Neural Network Layers Without Attention}

\begin{figure}
  \centering
  \includegraphics[width=0.95\linewidth]{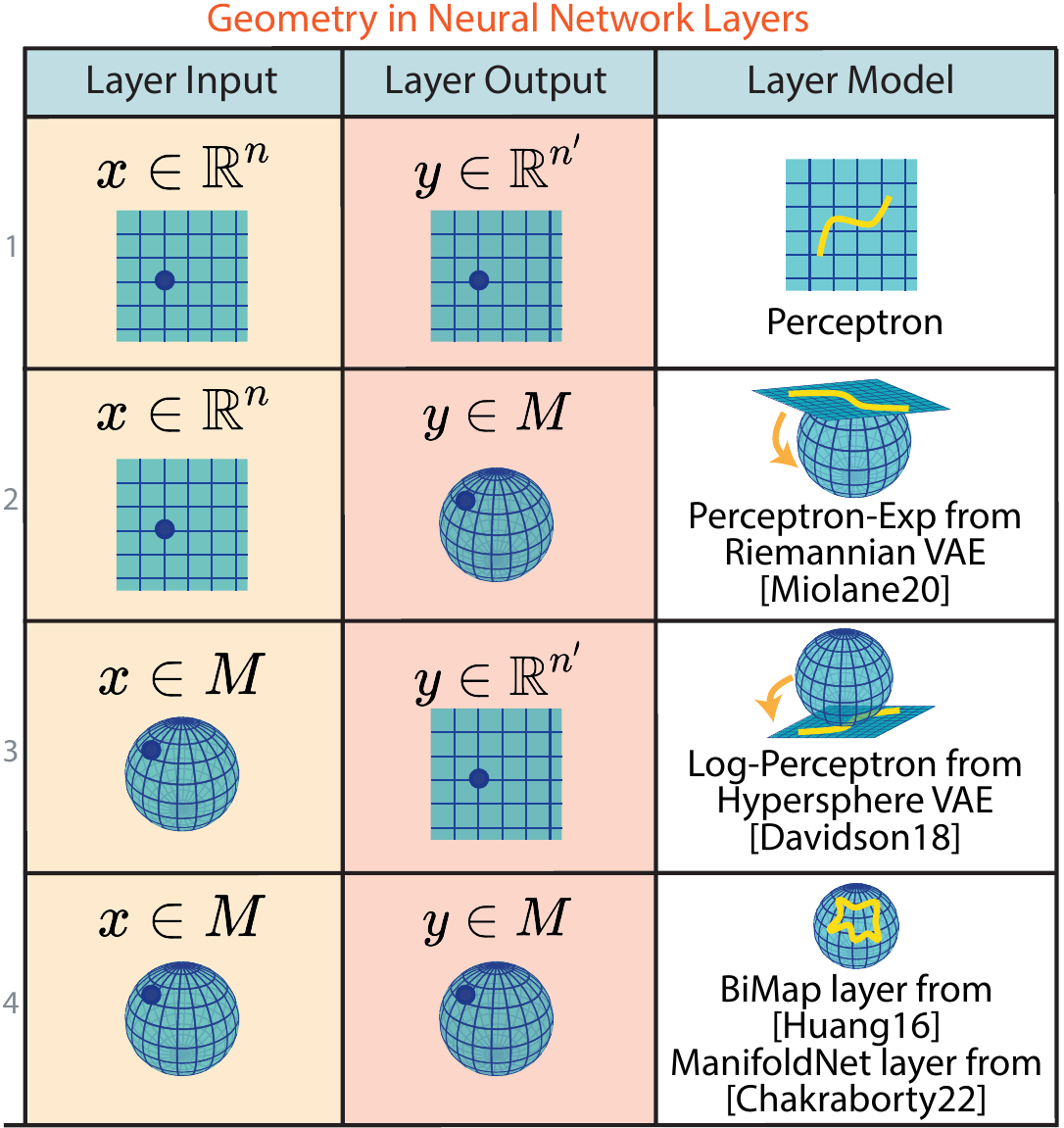}
  \caption{\textbf{Geometry in Neural Network Layers} organized according to the mathematical properties of the layer inputs $x$ and outputs $y$. Inputs and outputs that are data represented as coordinates in a space, typically, a Euclidean space $\mathbb{R}^n$ or a manifold $M$. Notations: $\mathbb{R}^n$: Euclidean space; $M$: Manifold.}
  \label{fig:geo-deep-learning}
\end{figure}

\begin{figure}
  \centering
  \includegraphics[width=0.95\linewidth]{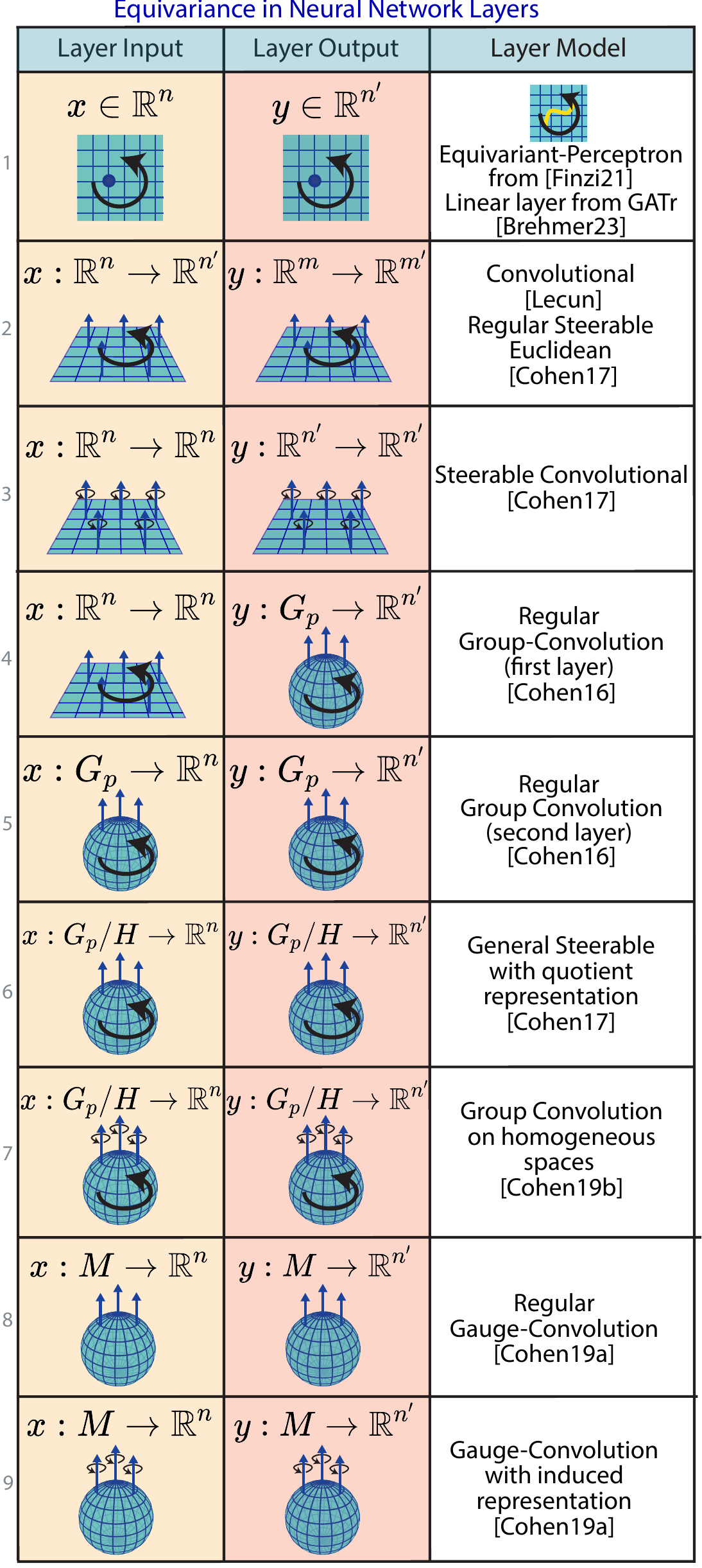}
  \caption{\textbf{Algebra in Neural Network Layers} organized according to the mathematical properties of the layer inputs $x$ and outputs $y$, which are both signals, i.e., functions from a domain to a codomain. The black curved arrows represent a group action on a space, such as a signal's domain. Notations: $\mathbb{R}^n$: Euclidean space; $M$: Manifold; $G_p$: group, $G_p/H$ Homogeneous manifold for the group $G_p$.}
  \label{fig:alg_deep_learning}
\end{figure}

\begin{figure}
  \centering
  \includegraphics[width=0.95\linewidth]{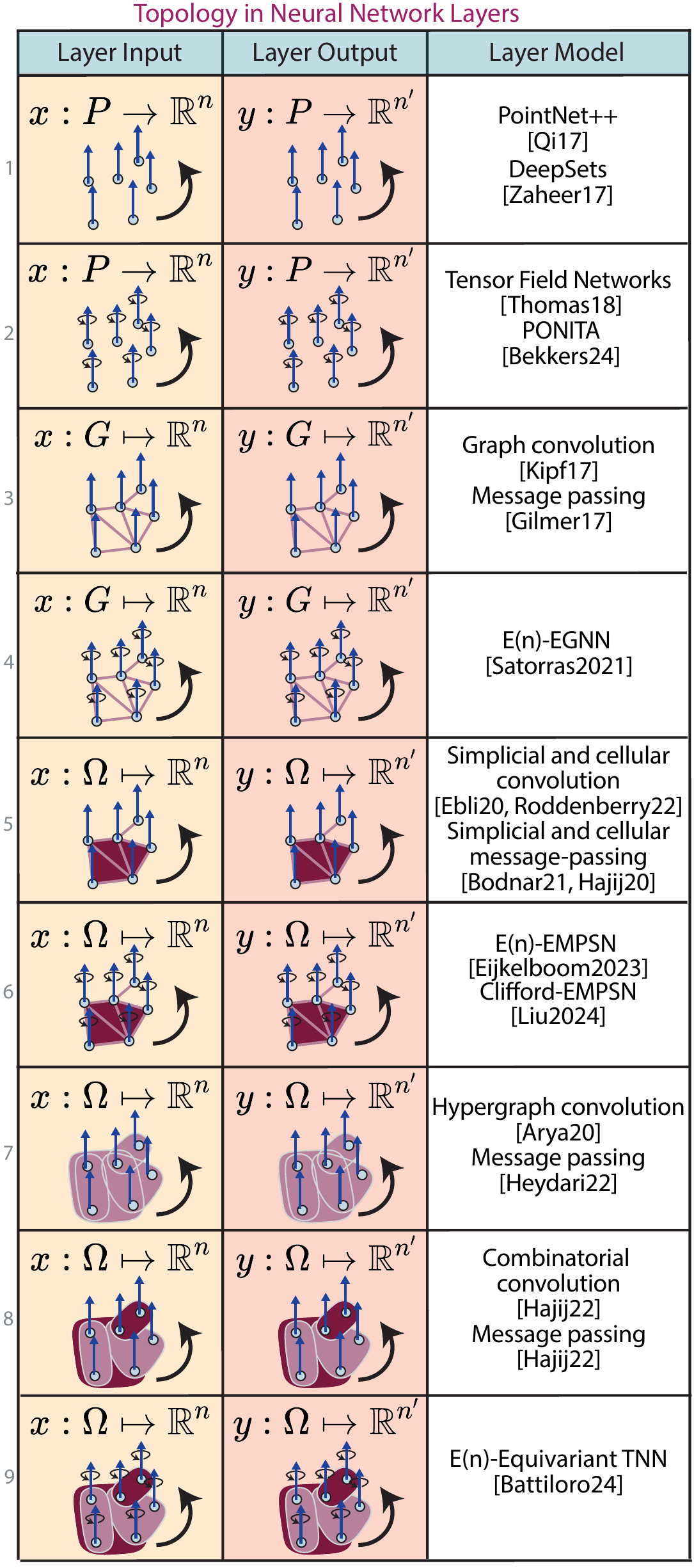}
  \caption{\textbf{Topology in Neural Network Layers} organized according to the mathematical properties of the layer inputs $x$ and outputs $y$, which are both signals, i.e., functions from a domain to a codomain. The black curved arrows represent a group action on a space. Notations: $\mathbb{R}^n$: Euclidean space; $P$: point set; $G$: graph; $\Omega$: topological space.}
  \label{fig:topo_deep_learning}
\end{figure}

Figures \ref{fig:geo-deep-learning}, \ref{fig:alg_deep_learning}, and \ref{fig:topo_deep_learning} organize deep learning methods into a taxonomy based on the mathematical properties of the input and output of neural network layers (first two columns), as well as on the properties of the layer model (third column). The rows show different types of layers that have been published in the literature or in preprints.

\vspace*{2mm}
\paragraph*{\textcolor{geometry}{Geometry in Neural Network Layers}}  In this section, we categorize neural network layers that consider their inputs and outputs as coordinates in spaces equipped with geometric structures.
The first category, layers with Euclidean input and Euclidean output, in row 1 of Fig. \ref{fig:geo-deep-learning} is exemplified by the Perceptron layer \citep{rosenblatt1958perceptron}. This foundational layer is commonly used as a component in a deep neural network comprised of several identical layers: the celebrated Multi-Layer Perceptron (MLP).

Next, we consider layers with Euclidean input with manifold output (row 2 of Fig. \ref{fig:geo-deep-learning}). The Perceptron-Exp layer \cite{miolane2020learning} extends the Perceptron layer to produce outputs on manifolds. Here, Exp denotes the Riemannian exponential map, which is applied to the result of the Perceptron. Indeed, Exp is an operation that maps tangent vectors to points on a manifold, i.e., that maps an input in an Euclidean space to an output on the manifold. Only the Perceptron component of this layer has learnable weights; the manifold needs to be known a priori in order to specify and implement the appropriate exponential map. Additionally, in general, there is no analytical expression for the Exp map, which needs to be computed numerically. To avoid this computational cost, this layer is best implemented for manifolds whose Exp enjoys an analytical expression. 

The third category, manifold input with Euclidean output, is represented by the Log-Perceptron layer \cite{Davidson2018HypersphericalAuto-encoders}  (row 3 of Fig. \ref{fig:geo-deep-learning}). This layer generalizes the Perceptron to accept inputs from manifolds, applying the Riemannian logarithm map (Log) prior to the Perceptron. The Log operation converts points on manifolds into tangent vectors, effectively serving as the inverse of the Exp map. Thus, it can be viewed as the inverse of the Perceptron-Exp layer \cite{miolane2020learning}. Here again, knowledge of the manifold is essential for implementation, and preference is given to manifolds with an analytically expressible Log.

Finally, we consider the manifold input with manifold output configuration (row 4 of Fig. \ref{fig:geo-deep-learning}). This is first illustrated by the Bimap and SPDNet layers from \citet{Huang2016ALearning}, which focus on symmetric positive definite (SPD) matrices constrained to the manifold of SPD matrices. More generally, ManifoldNet \citep{chakraborty2020manifoldnet} builds on the reformulation of convolutions as weighted Fréchet means of \citet{Pennec2006b} to propose a layer whose inputs and outputs are both coordinates on a manifold. In this layer, a weighted mean of the inputs is computed, where the weights are learned with classical backpropagation.  
We note that, in general, there is no analytical expression for the weighted Fréchet mean, which needs to be obtained by optimization. To avoid this computational cost, approximations using tangent means are also considered.

\vspace*{2mm}

\paragraph*{\textcolor{algebra}{Algebra in Neural Network Layers}} Here, we consider neural network layers used in \textit{equivariant deep learning}, which leverages the concepts of symmetries and group actions. We refer the reader to \citet{cohen2021equivariant} and \citet{weiler2024equivariant} for a detailed study of this field, and to \citet{kondor2018generalization} for foundational theoretical results on equivariance with respect to any compact group. This line of work builds the symmetries natural to a data domain, such as rotation symmetries for image classification, into the structure of the model. This facilitates weight sharing across different transformations so that the same convolutional filter can be used to detect a given feature in an image in, for example, all orientations. To accomplish this, the input and output spaces of these layers are equipped with group actions, and the layers are defined to be compatible with these actions. Intuitively, if a layer is \textit{equivariant} to a group action, this means that, if the layer produces an output $y$ for a given input $x$, then it should also produce a transformed $y$ for any transformed $x$\textemdash where transformed here means ``acted upon by the same group element.'' \footnote{{\color{black} We note that several authors have proposed categorical deep learning, a broader framework that generalizes equivariant deep learning \citep{gavranovic2024position}. However, due to the infancy of this field, a detailed discussion lies beyond the scope of this review.}}

Fig. \ref{fig:alg_deep_learning} highlights key examples of equivariant layers. The first row shows a layer that considers Euclidean input and output coordinates equipped with group action. The Equivariant-Perceptron layer from \citet{finzi2021} and the linear layer in the Geometric Algebra Transformer (GATr) \cite{brehmer2023geometric} illustrate this configuration. The work by \citep{finzi2021} allows us to build an Equivariant-Perceptron layer given any matrix group action on inputs and outputs. Next, we consider the same layer setup except with data as signals in space, instead of coordinates (row 2). In this case, the input and output could be images or feature maps, defined over the domains $\mathbb{R}^n$ and $\mathbb{R}^m$, respectively. The classic convolutional layer \citep{LeCun1998Gradient-BasedRecognition} is a layer with translation equivariance: a translation of the input image $x$ yields a translated feature map $y$ as output. Regular steerable convolutional layers \citep{cohen2017steerable} generalize this approach beyond the group of translations. Here, the adjective ``regular'' means the group action is only on the domain of the input and output signals. By contrast, row 3 shows layers for which both the domain and codomain of each signal are equipped with a group action, such as the Steerable Convolutional layer \citep{cohen2017steerable}.

A different class of layers considers signals defined over Lie groups $G_p$, where we recall that a Lie group is a manifold that is also a group. Specifically, the first layer of the Group Convolutional Network \cite{cohen16} (row 4) takes as input a Euclidean signal, typically an image, and outputs a signal equipped with group action, $y : G_p/H \mapsto \mathbb{R}^n$. The second layer of this network (row 5) takes this signal as an input and outputs a new signal equipped with group action. The group action here is the group composition.

Next, layers in rows 6 and 7 consider  signals of the form $x, y : G_p/H \mapsto \mathbb{R}^n$. Here, the domain $G_p/H$ defines a so-called homogeneous manifold. This is a manifold equipped with a group action  \textcolor{black}{such that}, for every pair of points on the manifold, there exists a group element that can transform one point onto the other via the action. For example, the two-sphere $S^2$ is homogeneous for the group of 3D rotations, leading to the introduction of spherical CNNs \citep{cohen2018spherical,kondor2018clebsch}. In Fig. \ref{fig:alg_deep_learning} row 6, General Steerable Convolutional Layer with the so-called quotient representation \citep{cohen2017steerable} falls in this category, equipping the domain with group action. In Fig. \ref{fig:alg_deep_learning} row 7, \citep{cohen2019general} further equips the Euclidean codomains of both input and output signals.

The last category of layers (Fig. \ref{fig:alg_deep_learning} rows 8, 9) considers input and output signals with manifold domains and Euclidean codomains. Signals are thus represented as $x, y : M \mapsto \mathbb{R}^n$. Here, the concept of gauge equivariance replaces the group equivariance. Gauge equivariance describes the idea of being agnostic to the orientation of a local coordinate system on the manifold of interest $M$. \citet{cohen2019gauge} proposes such a channel-wise gauge convolutional layer.  \textcolor{black}{Specifically, the layer is defined such that for a given input $x$ and output $y$, a different choice of local coordinate system on the input signal's manifold, i.e. its gauge, will yield an equally transformed output. Unlike the global transformation imposed by a group action, such as rotation, these layers deal with preserving local transformations.} These layers can be generalized to consider group actions on the codomains of the signals, as does the general gauge convolutional layer with induced representation \citep{cohen2019gauge}.

\vspace*{2mm}

    \paragraph*{\textcolor{topology}{Topology in Neural Network Layers}} Here, we categorize neural network layers that consider their inputs and outputs as signals over a topological domain. We refer the reader to the survey by \citet{papillon2023architectures} for a comprehensive overview of the neural network layers within this category, and we focus only on  \textcolor{black}{illustrative examples}. These layers are all equivariant to the permutation of the elements of their topological domains: permutation of the points in a set, of the nodes in a graph, etc. As such, all these layers are, at a minimum, equipped with group action on their domain.

    We begin with layers that consider Euclidean signal on a set, depicted in the first row of Fig. \ref{fig:topo_deep_learning}. PointNet \citep{Qi_2017_CVPR} processes point clouds, which have signal of the form $x, y : P \mapsto \mathbb{R}^n$ where $P$ is a set. The model is equivariant to a permutation in $P$ through its use of a symmetric function (max pooling). PointNet++ \citep{qi2017pointnetpp} operates on the same signal, using a series of PointNet layers to learn patterns at different scales by imbuing hierarchy within the point cloud. In both of these works, the experiments are restricted to run only on point clouds in 2D or 3D. \textcolor{black}{Deep Sets \citep{deepsets} characterizes what is necessary and sufficient in terms of parameter sharing for a layer to be equivariant (two examples are summation and max pooling).}
We can extend these layers to also equip the signal's codomain with group action (row 2), instead of just the domain. Tensor Field Networks (TFN) \citep{thomas2018tensor}, and PONITA \citep{bekkers2024fast} process signals in this category. Like PointNet++, TFN and PONITA process 2D and 3D point clouds and are equivariant to the permutation of their points in $P$. TFN and PONITA are additionally equivariant to 3D rotations and translations.
    
    Going beyond a simple set, the next set of layers considers signals defined over graphs (Fig. \ref{fig:topo_deep_learning} rows 3, 4). These layers process signals of the form $ x, y : G \mapsto \mathbb{R}^n $, i.e., defined over a graph $G$. This category is exemplified by the Graph Convolutional layer \citep{kipf2017semisupervised} and Message Passing layers \citep{gilmer17messagepassing}. Such layers can be generalized to also consider a group action on the codomain of the signals (row 4). We find, for example, the $E(n$-Equivariant Graph Neural Networks (EGNN) by \citet{satorras2021n}.

What if the underlying topology of the data is more accurately represented by multi-way relationships, rather than the pairwise connections of graphs? The following works address this by leveraging richer topological spaces, denoted as $\Omega$, as their signal's domain. Rows 5 and 6 show layers that process signals defined over simplicial and cellular complexes, which both allow for hierarchical relationships, i.e. faces, between edges. Examples include simplicial convolutional \citep{ebli2020simplicial}, cellular convolutional \citep{roddenberry22}, simplicial message passing \citep{bodnar2021mpsn}, and cellular message passing \citep{hajij2020cell} layers. In the case where group action also equips the co-domain (row 6), we find, for example, the $E(n)$-Equivariant Message Passing Simplicial Networks (EMPSN) \citep{eijkelboom2023n} and the Clifford group Equivariant Message Passing Simplicial Networks (EMPSN) \citep{liu2024clifford}. These both introduce group actions on the codomain, but for different groups.

    In the same spirit, row 7 describes layers that process signals defined over the hypergraph domain, which allows for multi-way edges, i.e., hyperedges, connecting many nodes at once. Examples include the Hypergraph Convolutional layer \citep{arya2020} and the Hypergraph Message Passing layer \citep{heydari22}. The last topological space that arises in the field is the combinatorial complex, which combines the hierarchical relationships of cellular complexes with the set-wise flexibility of hypergraphs (row 8). Examples of neural network layers that process this type of signals are the Combinatorial Convolutions \citep{hajijtopological} and Combinatorial Message Passing \citep{hajijtopological}. We can further leverage algebraic structure by equipping the codomain with a group action (row 9). The $E(n)$-equivariant topological neural networks from \citet{battiloro2024n} are an example of this configuration. The layers of this network process geometric features in the codomain $\mathbb{R}^n$, such as velocities or positions associated with elements of $\Omega$.

\subsection{Neural Network Layers with Attention} \label{sec:attention_mech}

We now review mathematical structures in neural network layers that leverage the attention mechanism, focusing on how non-Euclidean structures can enrich the structure of the attention coefficients and the attention layers.  \textcolor{black}{Attention mechanisms emerged as a transformative approach with foundational works by \citet{graves2014neural, graves2016hybrid} introducing dot-product attention, and \citet{bahdanau2015neural} applying it to machine translation. It was later widely popularized by the Transformer architecture \citep{vaswani2017transformer}. At its core, attention addresses the fundamental challenge of selectively focusing on relevant parts of input data while creating direct connections between arbitrary positions, thereby overcoming the limitations of traditional architectures in handling long-range dependencies. We devote a section in our study to attention mechanisms because they represent a distinct and increasingly dominant paradigm in neural network design, with unique mathematical properties that complement traditional approaches in our taxonomy. }

\begin{figure}
  \centering
  \includegraphics[width=1.0\linewidth]{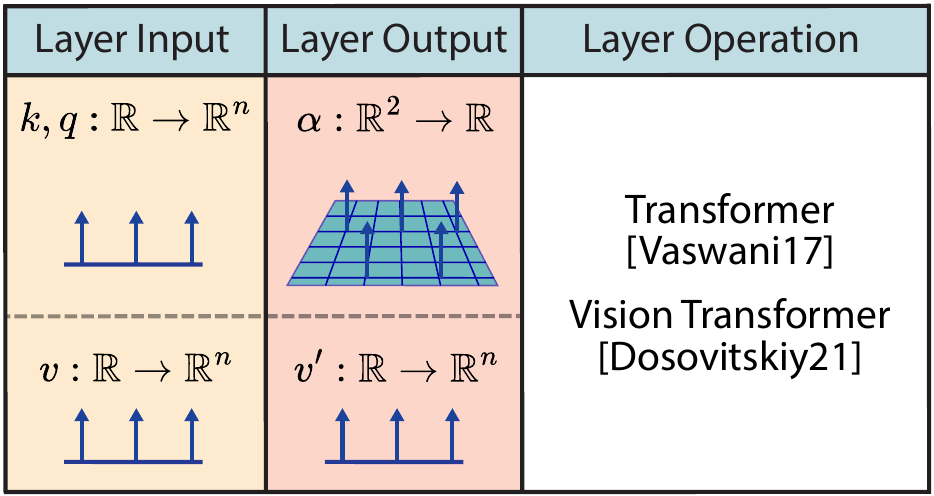}
  \caption{ \textcolor{black}{\textbf{Classical Transformers} illustrated by the mathematical properties of the attention coefficients and of the attention layer. The key $k$ and the query $q$ are the inputs to the attention coefficient $\alpha$; the value $v$ is the input to the attention layer, and the output value $v'$is the weighted result of that layer. Inputs and outputs are represented as signals, i.e., as functions from a domain to a codomain. Notation: $\mathbb{R}^n$: Euclidean space.}}
  \label{fig:trad_transformer}
\end{figure}

Let us briefly unpack how attention works. In a layer, the attention coefficient $\alpha$ is computed from a query $q$ and a key $k$. Hence, we examine the structure of the key $k$ and the query $q$ inputs, represented as signals over a domain. \textcolor{black}{For example, the traditional transformer \citep{vaswani2017transformer}, depicted in Fig. \ref{fig:trad_transformer}, considers keys and queries as functions over a one-dimensional domain $\mathbb{R}$ representing time.} The output attention coefficient $\alpha$ is represented as a signal over the product domain: in the transformer example, it is a function over the product domain $\mathbb{R} \times \mathbb{R}$. Second, the attention layer transforms input values $v$ to output values $v'$ through the attention coefficients $\alpha$. Hence, we look at the mathematical properties of $v$ and $v'$, both represented as signals over a domain. In the classical transformers, they are signals over the time domain $\mathbb{R}$.

 \textcolor{black}{Another example of a classical transformer is the Vision Transformer~\citep{dosovitskiy2021image}, which divides an image into a sequence of image patches: hence, patches in $\mathbb{R}^n$ over $\mathbb{R}$. We highlight the difference between the mathematical representation of these signals, and their computational representation during an actual implementation of the transformer. Mathematically, we represent the signals' domain as the continuous real line $\mathbb{R}$. Computationally, however, this real line is discretized into $T$ steps and associated $T$ discrete tokens (for either words or image patches). Yet, the representation as the real line is useful to unify the original transformer with the more complicated layers introduced next.}

\begin{figure}
  \centering
  \includegraphics[width=1.0\linewidth]{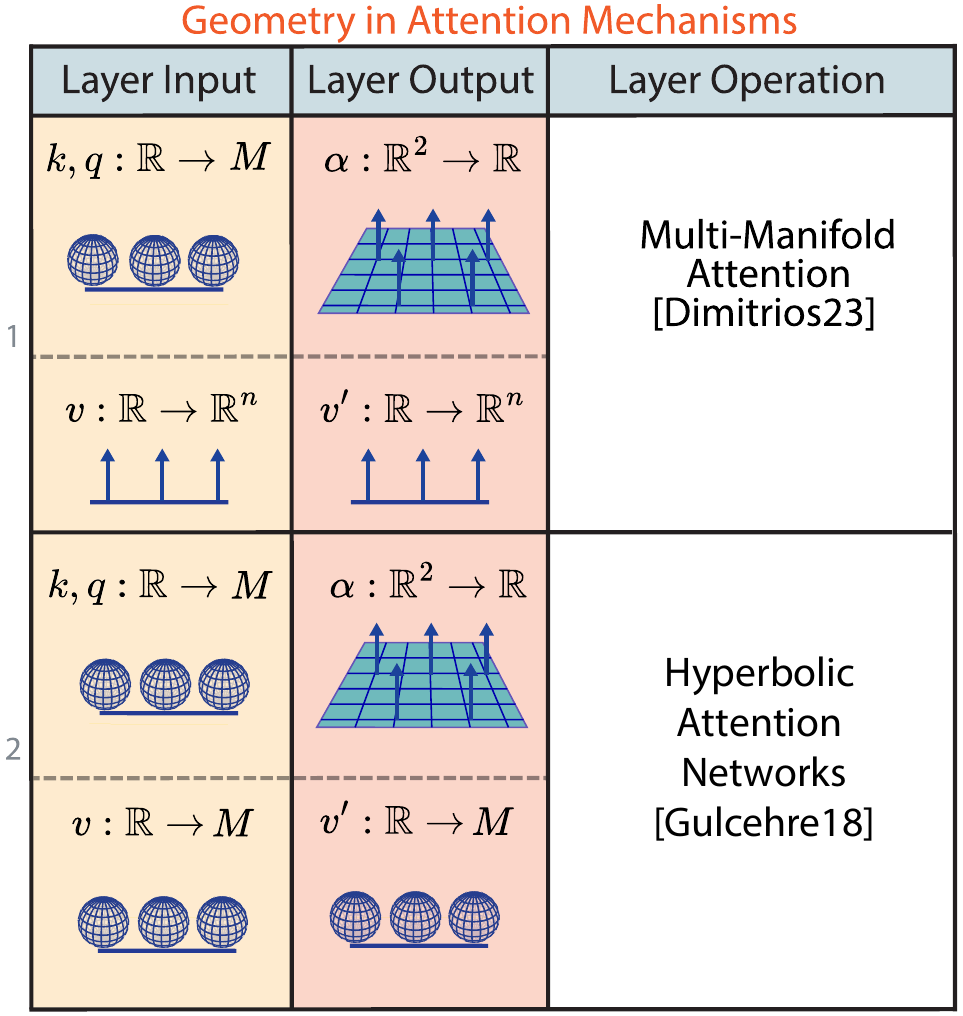}
  \caption{\textbf{Geometry in Attention Mechanisms} categorized according to the mathematical properties of the attention coefficients (first subrow of each row) and of the attention layer (second subrow of each row). The key $k$ and the query $q$ are the inputs to the attention coefficient $\alpha$; the value $v$ is the input to the attention layer, and the output value $v'$is the weighted result of that layer. Inputs and outputs are represented as signals, i.e., as functions from a domain to a codomain. Notations: $\mathbb{R}^n$: Euclidean space; $M$: Manifold.}
  \label{fig:geo_transformers}
\end{figure}

\begin{figure}
  \centering
  \includegraphics[width=1.0\linewidth]{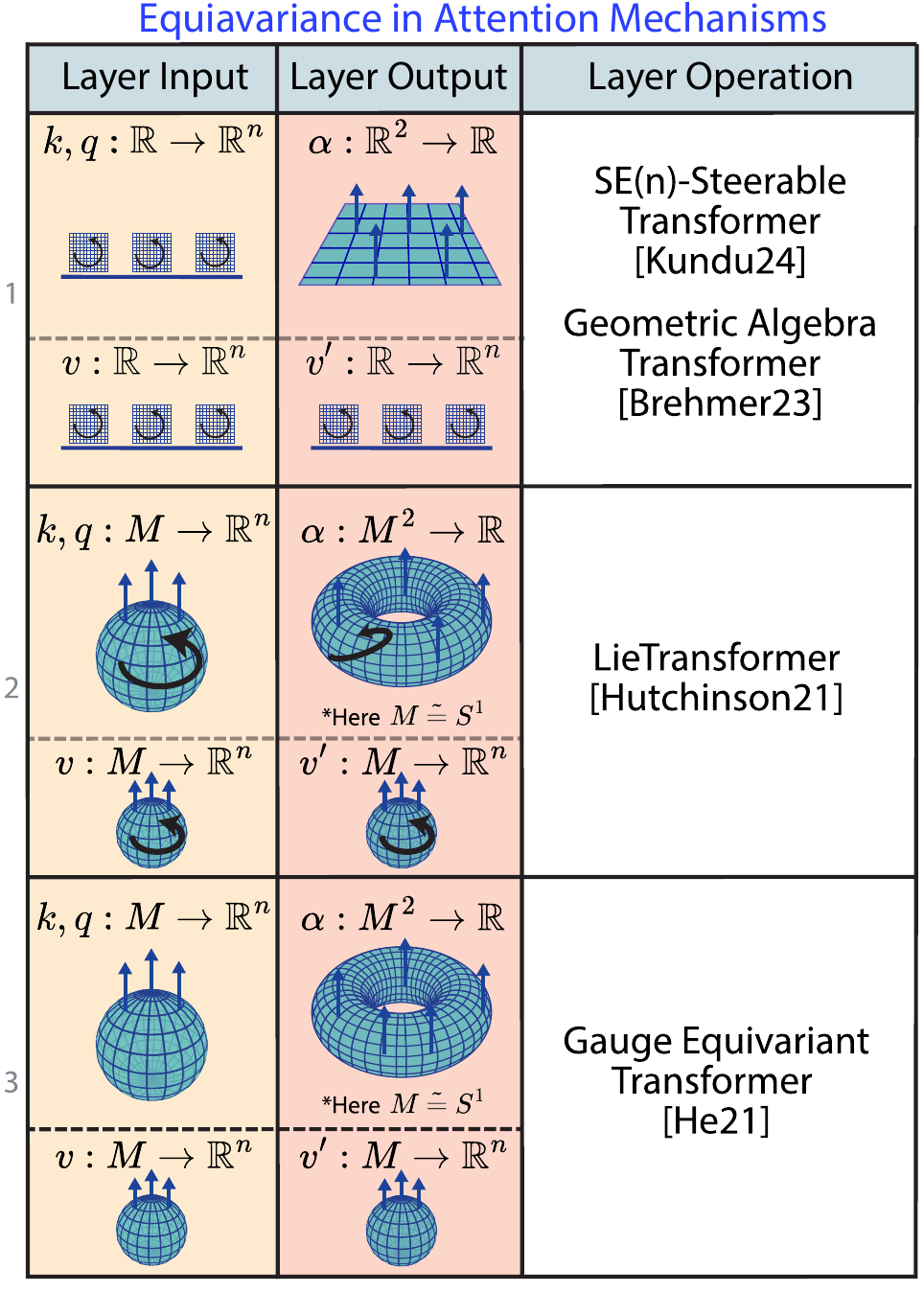}
  \caption{\textbf{Algebra in Attention Mechanisms} categorized according to the mathematical properties of the attention coefficients (first subrow of each row) and of the attention layer (second subrow of each row). The black curved arrows represent the action of a group on a signal's domain or codomain. Notations: $\mathbb{R}^n$: Euclidean space; $M$: Manifold.}
  \label{fig:alg_transformers}
\end{figure}

\begin{figure}
  \centering
  \includegraphics[width=0.85\linewidth]{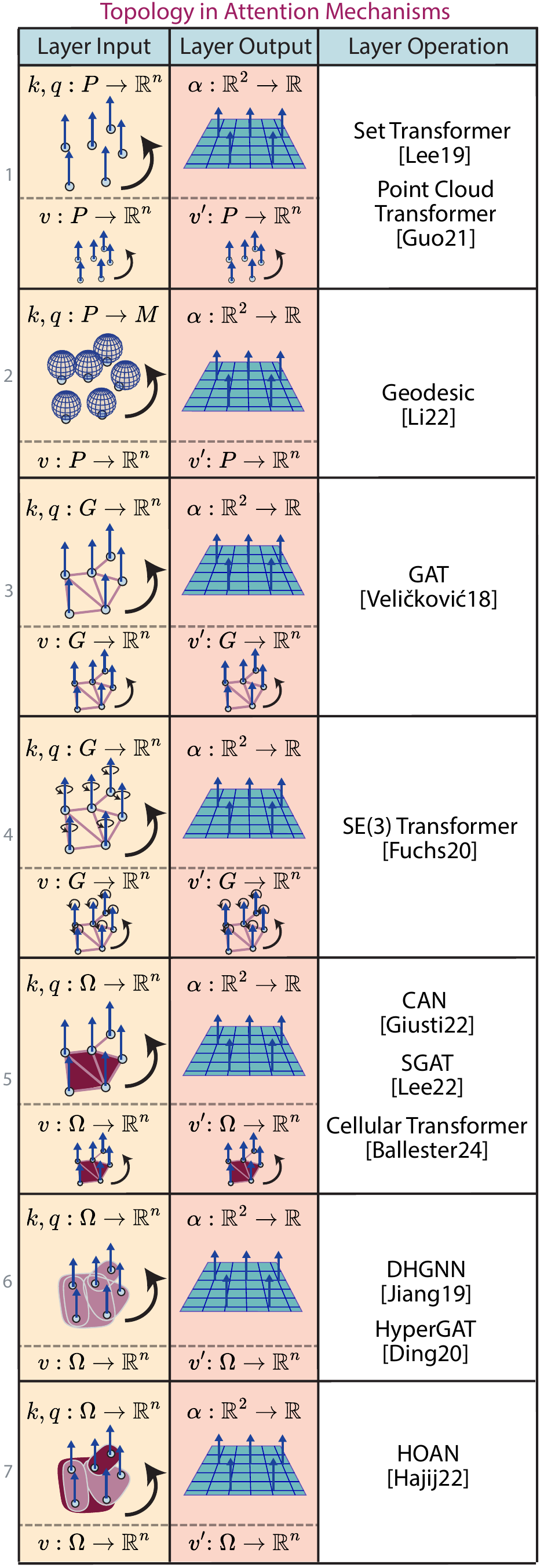}
  \caption{\textbf{Topology in Attention Mechanisms} categorized according to the mathematical properties of the attention coefficients and of the attention layer. Notations: $\mathbb{R}^n$: Euclidean space; $M$: Manifold; $P$: point set; $G$: graph; $\Omega$: topological space.}
  \label{fig:topo_transformers}
\end{figure}

 \textcolor{black}{With the classical case in mind,} we now turn to Figures~\ref{fig:geo_transformers}, \ref{fig:alg_transformers}, and \ref{fig:topo_transformers}, which organize attention coefficients and layers into a taxonomy based on the mathematical properties of their non-Euclidean inputs and outputs. Each row highlights the properties of the domains and codomains of the key $k$, query $q$, coefficient $\alpha$, input value $v$ and output value $v'$ signals.

\vspace*{2mm}
\paragraph*{\textcolor{geometry}{Geometry in Attention Mechanisms}} We consider the case where keys and queries are defined as manifold signals on a Euclidean domain (Fig. \ref{fig:geo_transformers} row 1). This row introduces geometry in the codomains of the keys and query signals, which write $k, q : \mathbb{R} \mapsto M$. Meanwhile, the attention coefficients $\alpha$, the input and output values $v, v'$ have the same structure as in the classic transformer. The multimanifold attention mechanisms by \cite{konstantinidis2023} illustrates this configuration. Specifically, this work considers the classical attention coefficient $\alpha$ as the computation of a Euclidean distance between key and query. Accordingly, their proposed geometric attention coefficient replaces the Euclidean distance by a Riemannian geodesic distance between key and query, which are interpreted as elements of a manifold: the manifold of SPD (symmetric positive definite) matrices, the Grassmann manifold, or both\textemdash hence the term ``multi''-manifold. We note that the input data to this transformer architecture is still Euclidean, since this attention mechanism is proposed for images in vision transformers. However, the way this data is processed by the transformer's internal layers is non-Euclidean.

\textcolor{black}{In Fig. \ref{fig:geo_transformers} row 2, we consider the case in which queries, keys and values are first mapped from Euclidean activations onto the hyperboloid manifold. The hyperbolic transformer \citep{gulcehre2018hyperbolic} computes attention weights via an exponential (or sigmoid) of the hyperbolic distance between $q_i$ and $k_j$, and aggregates Klein-model values with the Einstein midpoint. Although raw inputs such as images, graph node features or word embeddings start in Euclidean space, all attention operations in the network occur in hyperbolic space.}

\vspace*{2mm}

\paragraph*{\textcolor{algebra}{Equivariance in Attention Mechanisms}} 

We first consider layers where keys, queries and values are defined as signals on Euclidean domains with group action on the codomain. Fig. \ref{fig:alg_transformers} row 1 illustrates an attention mechanism that includes additional algebraic structure in both keys, queries, input and output values. The Geometric Algebra Transformer represents this configuration~\cite{brehmer2023geometric}. Its layers were designed to process ``geometric data'' defined as scalars, vectors, lines, planes, objects and their transformations (e.g., rotations) in 3D space. Such geometric data is encoded into multivectors, which are elements of the projective geometric algebra also called the Clifford algebra. For simplicity, we consider this space as a Euclidean space $\mathbb{R}^n$ with algebraic structure. The attention coefficients are group invariant, while the attention layer is group equivariant, for the Euclidean group $E(3)$ of translations, rotations and reflections in 3D space. This configuration is also illustrated by the Steerable Transformer \citep{kundu2024steerable} which processes Euclidean codomains $\mathbb{R}^n$ with equivariance to the special Euclidean group $SE(n)$, for application to image processing and machine learning tasks.

In Fig. \ref{fig:alg_transformers} row 2, we generalize row 1 to signals defined over manifold domains. As before, this row also introduces geometric structure in the keys, queries, values. In contrast to the previous row, however, this layer brings geometry into the domains of the signals: $k, q, v, v'$ whereas the above layer brought geometry in their codomains. Consequently, for this row, the attention coefficients $\alpha : M \times M \mapsto \mathbb{R}$ have the product manifold $M \times M$ as their domains. This configuration is illustrated in the Lie Transformer \citep{hutchinson2020lietransformer}, where the manifolds of interest are Lie groups and their subgroups. We note that the data processed by this architecture does not have to belong to a Lie group; only to be acted upon by a Lie group. A lifting layer is introduced to convert the raw data into Lie group elements, which are then handled by the Lie Transformer. The attention layer is then equivariant.

Lastly, we explore attentional layers that feature built-in gauge equivariance and invariance (Fig. \ref{fig:alg_transformers} row 3), rather than the group equivariance from before. The Gauge invariant transformer in \cite{he2021gaugeinvariant} presents such a configuration: the attention coefficients are gauge invariant, and the attention layer is gauge equivariant. This work exclusively focuses on two-dimensional manifolds $M$ embedded in 3D Euclidean space.

\vspace*{2mm}

\paragraph*{\textcolor{topology}{Topology in Attention Mechanisms}} We now turn to attentional layers defined on topological spaces. All of these layers are permutation equivariant, meaning they respect domain-level group actions. Their key differences lie in how they define the underlying topological domain on which the data resides. Following the same approach as in the non-attentional case, we will present these domains in roughly increasing order of topological complexity, starting with the simple set.

Fig. \ref{fig:topo_transformers} row 1 introduces attentional layers that define Euclidean signals on the set domain. The Set Transformer (ST) \citep{lee2019st} and the Point Cloud Transformer (PCT) \citep{guo2021pct} are two examples of this configuration. These layers' equivariance to the group of permutations means they are equivariant to any permutation of the points' indexing in the set (resp., in the point cloud). Going beyond simple Euclidean features, certain set-based layers (Fig. \ref{fig:topo_transformers} row 2) leverage manifold signals, meaning these layers
include geometric structure on top of the set-based topological structure. Specifically, they use manifold codomains for keys $k$, queries $q$ and Euclidean signals for $v$ and $v'$, all on set domains. The geodesic transformer \citep{li2022geodesictransformer} provides an architecture that processes this configuration. Similar to the Multi-Manifold Attention of Fig. \ref{fig:geo_transformers} row 1, the attention coefficient of the geodesic transformer is computed using a geodesic distance between keys and queries: either a graph-based geodesic distance, \textcolor{black}{or a Riemannian geodesic distance on an oblique manifold. The latter manifold refers to the set of matrices whose columns are unit norm but not necessarily orthogonal. It is a useful choice because the unit norm constraint helps stabilize optimization, preventing issues like exploding or vanishing gradients, while offering more flexibility than manifolds that enforce full orthogonality.}

Slightly increasing topological complexity, Fig. \ref{fig:topo_transformers} rows 3 and 4 describe layers defined on the graph domain. The most well-known example is the Graph Attention Transformer (GAT)  \citep{velickovic2018graph}, which features the same group action (permutation) equivariance on the domain (row 3). We can additionally equip the codomain of a graph attentional layer with group action (row 4), as is the case in the $SE(3)$-Transformer~\citep{fuchs2020se3}. Here, the codomain of the signals is additionally restricted to $\mathbb{R}^3$, equipped with an action of the group of translations and rotations in 3D $SE(3)$. This transformer was proposed to process 3D point clouds, and provides $SE(3)$-invariant attention coefficients and $SE(3)$-equivariant attention layer.

Going beyond pairwise relations, the remaining rows of Fig. \ref{fig:topo_transformers} consider the same richer domains as outlined in Fig. \ref{fig:topo_deep_learning} rows 5-9 and detailed in the survey \citet{papillon2023architectures}. While they all feature group action (permutation) equivariance on the domain, none extend algebraic structure to the codomain, as is the case for graphs (Fig. \ref{fig:topo_transformers} row 4). We briefly detail the works that exemplify this line of work: Simplicial Graph Attention Network (SGAT) \citep{lee2022sgat} and Cell Attention Networks (CAN) \citep{giusti2023} define their signals on simplicial and cellular complexes, respectively. The Cellular Transformer \citep{ballester2024attending} additionally includes positional encodings. Dynamic Hypergraph Neural Networks (DHGNN) \citep{jiang2019dhgnn} and Hypergraph Attention Networks (Hyper GAT) \citep{ding2020hypergat} leverage hypergrpahs. The Higher Order Attention Network (HOAN) architecture \citep{hajijtopological} uses the combinatorial complex domain, which combines the relations featured in cellular complexes and hypergraphs.

This concludes the review of non-Euclidean in deep neural network layers. While a great diversity of layers and mechanisms have been proposed, we hope that our illustrated taxonomy aids researchers in understanding the landscape and identifying opportunities for innovation and application. In the next sections, we turn to practical aspects of deploying topological and geometric machine learning methods in applications, including a table of common benchmarks used in the literature, a table of non-euclidean software libraries, and a review of key domains in which these approaches have been applied.

\section{Non-Euclidean Deep Learning Benchmarks}
\label{sec:benchmarks}
Here, we present a brief review of the benchmarks that have been considered in the non-Euclidean deep learning literature, compiling results from a broad sample of neural networks with topological, geometric, and  algebraic layers in Table \ref{tab:apps}, and highlighting the diversity of tasks and datasets used in the literature. 

\vspace{0.4cm}

\paragraph*{Tasks and Datasets} We first observe that a wide variety of task and benchmark datasets have been used in the literature, with little overlap between models. In other words, it is rare that two different models have been benchmarked on the same dataset. This is not surprising, since different models use different geometric, topological, and algebraic structures and different structures are well suited for different tasks.

There are, however several benchmarks that appear across models: MNIST and CIFAR for image classification, and Cora, Citeseer, and Pubmed for graph classification. Many geometrical models are tested by examining how well they model dynamical or physical systems. These results are not easily comparable across models, as the tasks are often customized for each paper. 

\begin{table*}
\footnotesize

\caption{\textbf{Applications and Benchmark of Neural Networks with Geometric, Topological and Algebraic Structures.}\\ We organize models according to whether it uses attention and their geometric, topological and algebraic structure, with the abbreviations: M: manifold, $G_p$: group, $S$: set, $G$: graph, $\Omega$: topological domain, $A$: algebra. Models are also organized based on which task they perform, and on which benchmark datasets. We include accuracies for benchmarks that two or more models use, converting test error to accuracy when needed, along with standard error if reported. Model parameters are listed if the paper reports them. N. R. means Not Reported. \label{tab:apps}}

\renewcommand{\arraystretch}{1.3}
\begin{tabularx}{\textwidth}{@{\extracolsep{\fill}} p{.005\textwidth}>{\raggedright}p{.17\textwidth}p{.02\textwidth}>{\raggedright}p{.25\textwidth}>{\raggedright}p{.30\textwidth}p{.03\textwidth}}

\hline
 & \textbf{Model} &  \multicolumn{1}{r}{\makebox[.02\textwidth]{\textbf{Structure}}} & \textbf{Task} & \textbf{Benchmark datasets} &  \multicolumn{1}{c}{\makebox[.02\textwidth]{\textbf{\# Params}}} \\

\hline
\multirow{25}{*}{\rotatebox[origin=c]{90}{\textbf{Without Attention}}}
  & Riemannian VAE \citelinktext{Miolane2019UnsupervisedNetworks}{Miolane19} & M & Dimension Reduction & Human Connectome Project (HCP) & N.R.\\
 & S-VAE/VGAE \citelinktext{Davidson2018HypersphericalAuto-encoders}{Davidson18} & M & Latent representation for image classification and link prediction & MNIST (93.4$\pm$ 0.2*), Cora (94.1$\pm$0.3), Citeseer (95.2$\pm$0.2), Pubmed (96.0$\pm$0.1) & N.R.\\
 & SPDNet \citelinktext{Huang2016ALearning}{Huang,VanGool16} & M & Visual classification (emotion, action, face) & AFEW, HDM05 and PaSC & N.R.\\
 & EMLP \citelinktext{finzi2021}{Finzi21} & $G_p$ & Dynamical modeling & Double pendulum & N.R.\\
 & LeNet-5 \citelinktext{LeCun1998Gradient-BasedRecognition}{LeCun98} & $G_p$ & Image classification & MNIST (99.2$\pm$0.1) & N.R.\\
 & Steerable CNN \citelinktext{cohen2017steerable}{Cohen,Welling17} & $G_p$ & Image classification & CIFAR (10: 76.3; 10+: 96.4; 100+: 81.2) & 4.4M\newline 9.1M\\
 & G-CNN \citelinktext{cohen16}{Cohen,Welling16} & $G_p$ & Image classification & Rotated MNIST, CIFAR (10: 93.5; 10+: 95.1) & 2.6M\\
 & G-CNN \citelinktext{cohen2019gauge}{Cohen19a} & $G_p$ & Climate, pointcloud segmentation & Climate Segmentation, Stanford 2D-3D-S & N.R. \\
 & E(n)-EGNN \citelinktext{satorras2021n}{Satorras2021} & $G_p$ & Molecular property prediction, dynamical modeling & QM9, N-body, Graph autoencoder & N.R. \\
 & PONITA \citelinktext{bekkers2024fast}{Bekkers2024} & $G_p$ & Molecular property prediction and generation, dynamical modeling & rMD17, QM9, N-body & N.R. \\
 & PointNet++ \citelinktext{qi2017pointnetpp}{Qi17} & S & Image, 3D, scene classification & MNIST (99.49), ModelNet40 (91.9), SHREC15, ScanNet & 1.7M\\
 & Tensor field network 
\citelinktext{Thomas2018TensorClouds}{Thomas18} & S & 3D-point-cloud prediction & QM9 & N.R.\\
 & GCN \citelinktext{kipf2017semisupervised}{Kipf,Welling17} & G & Link prediction & Cora, Citeseer, Pubmed, NELL & N.R. \\
 & enn-s2s \citelinktext{gilmer17messagepassing}{Gilmer17} & G & Molecular property prediction & QM9 & N.R.\\
 & SNN \citelinktext{ebli2020simplicial}{Ebli20} & $\Omega$ & Coauthorship prediction & Semantic Scholar Open Research Corpus & N.R.\\
 & MPSN \citelinktext{bodnar2021mpsn}{Bodnar21} & $\Omega$ & Trajectory, graph classification & TUDataset & N.R.\\
 & CXN \citelinktext{hajij2020cell}{Hajij20} & $\Omega$ & - & - & N.R.\\
 & HMPNN \citelinktext{heydari22}{Heydari22} & $\Omega$ & Citation node classification & Cora (92.2) & N.R.\\
 & CCNN \citelinktext{hajijtopological}{Hajij23} & $\Omega$ & Image segmentation, image, mesh, graph classification & Human Body, COSEG, SHREC11 & N.R.\\
 & E(n)-EMPSN \citelinktext{eijkelboom2023n}{Eijkelboom2023} & $\Omega$ & Molecular property prediction, dynamical modeling & QM9, N-body & 200K \\
 & Clifford-EMPSN \citelinktext{liu2024clifford}{Liu2024} & $\Omega$ & Pose estimation, dynamical modeling & CMU MoCap, MD17 & 200K \\
 & E(n) Equivariant TNN \citelinktext{battiloro2024n}{Battiloro2024} & $\Omega$ & Molecular property, air pollution prediction & QM9, Air Pollution Downscaling & 1.5M \\
\hline
\multirow{18}{*}{\rotatebox[origin=c]{90}{\textbf{With Attention}}}
 & Transformer \citelinktext{vaswani2017transformer}{Vaswani17} & - & Machine translation & WMT 2014 & N.R.\\
 & MMA ViT \citelinktext{konstantinidis2023}{Konstantinidis23} & M & Image classification, segmentation & CIFAR (10: 94.7, 100+: 77.5), T-ImageNet, ImageNet, ADE20K & 3.9M\\
  & GATr \citelinktext{brehmer2023geometric}{Brehmer23} & A & Dynamical modeling & N-body, artery stress, diffusion robotics & 4.0M\\
  & Steerable Transformer \citelinktext{kundu2024steerable}{Kundu2024} & $G_p$ & Point-cloud, Image classification & Rotated MNIST (99.03), ModelNet10 (90.4) & 0.9M \\
 & Lie Transformer \citelinktext{hutchinson2020lietransformer}{Hutchinson20} & $G_p$ & Regression, dynamics & QM9, ODE spring simulation & 0.9M \\
 & GET \citelinktext{he2021gaugeinvariant}{He21} & M & Shape classification, segmentation & SHREC07, Human Body Segmentation & 0.15M\\
 & Set Transformer \citelinktext{lee2019st}{Lee19} & S & Max value regression, clustering & Omniglot, CIFAR (100: 0.92$\pm$0.01 & N.R.\\
 & PCT \citelinktext{guo2021pct}{Guo21} & S & Point-cloud classification, regression, segmentation & ModelNet40 (93.2), ShapeNet (86.4), S3DIS & 1.4M\\
 & GSA \citelinktext{li2022geodesictransformer}{Li22} & S & Object classification, segmentation & ModelNet40 (93.3), ScanObjectNN, ShapeNet (85.9) & 18.5M\\
 & SE(3)-Transformer \citelinktext{fuchs2020se3}{Fuchs20} & G & Dynamics, classification, regression & N-body, ScanObjectNN, QM9 & N.R.\\
 & GAT \citelinktext{velickovic2018graph}{Veličković18} & G & Link prediction & Cora, Citeseer, Pubmed, PPI & N.R.\\
 & CAN \citelinktext{giusti2023}{Giusti23} & $\Omega$ & Graph classification & TUDataset & N.R.\\
 & Cellular Transformer \citelinktext{ballester2024attending}{Ballester24} & $\Omega$ & Graph classifical, Graph regression & GCB, Zinc, Ogbg Molhiv & N.R.\\
 & SGAT \citelinktext{lee2022sgat}{Lee22} & $\Omega$ & Node classification & DBLP$^2$, ACM, IMDB & N.R.\\
 & DHGNN \citelinktext{jiang2019dhgnn}{Jiang19} & $\Omega$ & Link, sentiment prediction & Cora (82.5), Microblog & 0.13M\\
 & HyperGAT \citelinktext{ding2020hypergat}{Ding20} & $\Omega$ & Text classification & 20NG, R8, R52, Ohsumed, MR & N.R.\\

\end{tabularx}
\end{table*}

\vspace{0.4cm}
\paragraph*{Number of parameters}

A key benefit of building mathematical structure into neural networks is that it constrains the hypothesis search space. If the structure is well matched to the problem, the model should require fewer parameters and fewer computations. Many papers mention this, but only a few report the number of parameters (see right column of Table~\ref{tab:apps}). As parameter and data efficiency are frequently cited as advantages of building structure into neural network models, we encourage authors to more regularly report parameter counts and computational cost in their papers along with performance metrics.

\footnotesize 

\begin{table*}[ht]
\footnotesize

\renewcommand{\arraystretch}{1.3}
\begin{tabularx}{\textwidth}{@{\extracolsep{\fill}} p{.18\linewidth}p{.28\linewidth}p{.4\linewidth}p{.05\linewidth}} 

\multicolumn{4}{c}{\textbf{\textcolor{geometry}{Geometry}}} \\\\
\hline
\textbf{Packages} & \textbf{Domains} & \textbf{Core Features} & \textbf{Stars}\\
\hline

GeomStats \citeyear{GEOMSTATS}
& Manifolds, Lie Groups, Fiber Bundles, Shape Spaces, Information Manifolds, Graphs
& Manifold operations, Algorithms, Statistics, Optimizers
& 1.3k
\\ 

GeoOpt \citeyear{geoopt2020kochurov}
& Manifolds
& Layers, Manifold operations, Stochastic optimizers for deep learning
& 917
\\ 

PyManOpt \citeyear{Townsend2016PymanoptAP}
& Manifolds, Lie Groups
& Manifold operations, Optimizers
& 816
\\ 

GeometricKernels \citeyear{mostowsky2024geometrickernels}
& Riemannian manifolds, graphs and meshes
& Gaussian process models
& 247
\\

PyRiemann \citeyear{pyriemann}
& SPD Matrices
& Machine Learning, Data Analysis for biosignals
& 676
\\

\\
\multicolumn{4}{c}{\textbf{\textcolor{topology}{Topology}}} \\\\
\hline
\textbf{Packages} & \textbf{Domains} & \textbf{Core Features} & \textbf{Stars}\\
\hline

PytorchGeometric \citeyear{Fey/Lenssen/2019}
& Graphs
& Baseline Models, Layers, Fast Basic Graph Operations, Datasets, Dataloaders
& 22.3k
\\ 

NetworkX \citeyear{hagberg2008exploring}
& Graphs, Digraphs, Multigraphs
& Data structures, Graph generators, Graph Algorithms, Network Analysis Measures
& 15.7k
\\  

DGL \citeyear{wang2019dgl}
& Graphs
& Baseline Models, Layers, Fast Basic Graph Operations, Datasets, Dataloaders, Framework-agnostic (PyTorch, Tensorflow, etc. are swappable)
& 13.9k
\\ 

DIG \citeyear{Liu2021DIGAT} 
& Graphs
& Baseline models, Datasets, Evaluation Metrics
& 1.9k
\\ 


AutoGL \citeyear{guan2021autogl}
& Graphs
& Neural Architecture Search, Hyper-Parameter Tuning, Ensembles
& 1.1k
\\ 

HyperNetX \citeyear{Praggastis2023HyperNetXAP}
& Hypergraphs
& Machine Learning Algorithms, Analysis, Visualization
& 606
\\ 

DHG \citeyear{gao2022hgnn}
& Graphs, hypergraphs, bipartite graphs, hypergraphs, directed hypergraphs, ...
& Models, Basic Operations, Dataloaders, Visualization, Auto ML, Metrics, Graph generators
& 737
\\ 

TopoModelX \citeyear{hajij2024topoxsuitepythonpackages}
& Graphs, colored hypergraphs, complexes
& Baseline Models, Layers, Higher-order message passing
& 205
\\ 

TopoNetX \citeyear{hajij2024topoxsuitepythonpackages}
& Graphs, colored hypergraphs, complexes 
& Topography Generators, Computing topological properties, Arbitrary cell attributes 
& 268
\\ 

TopoEmbedX \citeyear{hajij2024topoxsuitepythonpackages}
& Graphs, colored hypergraphs, complexes
& Representation learning, embeddings
& 80
\\ 

TopoBench \citeyear{topobenchmarkx2024}
& Graphs, hypergraphs, complexes
& Benchmarks, lifting, dataloaders, losses, training framework
& 116
\\

XGI \citeyear{Landry_XGI_2023}
& Hypergraphs, directed hypergraphs, symplical complexes
& Graph generators, metrics, algorithms, dataloaders, visualization
& 209
\\

\\
\multicolumn{4}{c}{\textbf{\textcolor{algebra}{Algebra}}} \\\\
\hline
\textbf{Packages} & \textbf{Domains} & \textbf{Core Features} & \textbf{Stars}\\
\hline
E3NN \citeyear{e3nn} & E(3) Equivariant Feature Fields & Group Convolutions, Steerable Group Convolutions & 1.1k \\ 
ESCNN \& E2CNN \citeyear{cesa2022a}
& E(n) Equivariant Feature Fields, Graphs
& Group Convolutions, Steerable Group Convolutions
& 584 \footnote{Stars inherited from e2cnn, which this extends.}
\\  
NequIP \citeyear{batzner2023}
& E(3) Equivariance on Graphs
& Group Convolutions, Steerable Group Convolutions
& 712
\\  

EMLP \citeyear{finzi2021}
& Matrix Groups, Tensors, Irreducible Representations, Induced Representations
& Programmatic generation of equivariant MLPs for arbitrary matrix groups in JAX
& 267
\\

PyQuaternion 
& Quaternions
& Quaternion operations, rotation representation conversions, differentiation, integration
& 357
\\ 
\end{tabularx}

\caption{\textbf{Software Packages for Machine Learning with Topology, Geometry, and Algebra.} We organize packages according to the mathematical, non-Euclidean structures they focus on. }
\label{table:software}

\end{table*}
\normalsize

\section{Non-Euclidean Software}\label{sec:software}

Table \ref{table:software} highlights publicly available software libraries that make the methods of this field computationally accessible. Here, we limit our discussion to libraries whose commit history suggests continued development and have a following indicated by at least 50 Github Stars. As shown by the number of stars and actively developed repositories, packages for topological methods are the most well developed, including important engineering foundations such as CUDA and C++ accelerated network primitives, and large collections of model implementations that continue to be maintained. The library ecosystem for geometric learning methods is quickly growing in interest and contributors, extending the packages beyond optimizers over specific manifolds to more general differential geometry tools. While the packages for algebra in machine learning are the most nascent, there have been exciting new developments within the past few years in making more specialized packages for accelerating group convolutions and other algebraic operations as the need for more specialized applications have emerged.

\section{Non-Euclidean Learning for Science}
\label{sec:apps}

Many problems in science and engineering are intrinsically non-Euclidean and thus provide an exciting opportunity for the application of non-Euclidean ML methods. Here, we briefly highlight key developments in selected application areas. We refer the reader to \citet{9046288, bronstein2021geometric, gaudelet2021utilizing, Rajpurkar2022AI, Li2022Graph, 10.1145/3535101, Wang2023ScientificDI} for more comprehensive discussions of applications.

\textbf{Chemistry and Drug Development}

Graph neural networks have become a workhorse for molecular analysis, treating molecules as graphs with atoms as nodes and bonds as edges \citep{gilmer17messagepassing, bronstein2021geometric} {\color{black}(Card C3 in Figure~\ref{fig:data-coord})}. Progress in this field has largely involved the construction of message-passing neural networks with favorable properties, such as equivariance to a growing family of group transformations {\color{black}(see examples of layers in Fig. \ref{fig:topo_deep_learning})}, novel forms of weight sharing, more expressive primitives, and more efficient formulations for parameterization and computation \citep{schutt2017schnet, thomas2018tensor, batzner2023, satorras2022en, bekkers2024fast}. Deep networks with geometric structure have also been used directly for drug screening to discover new antibiotics \citep{Stokes2020ADL}.

Recently, deep equivariant generative modeling has emerged as a powerful framework for molecule synthesis. Prior work by \citet{gebauer2020symmetryadapted, simonovsky2018graphvae, simm2020symmetryaware} establishes the importance of leveraging geometric properties for the synthesis of molecules. \citet{hoogeboom2022equivariant} introduces equivariant denoising diffusion models for molecule generation by directly generating 3D atomic coordinates, demonstrating improved quality and efficiency. This was recently extended by \citet{xu2023geometric} to perform equivariant diffusion over a molecular latent space, and by \citet{vignac2023midi} which achieves much higher stability for generated molecules on the GEOM-DRUGS dataset. Another line of work generates molecular invariants such as angles and distances, which are then used to produce coordinates \citep{Luo2022AnAF}. Recent work has also demonstrated the importance of equivariances and invariances for molecular conformer generation \citep{xu2022geodiff, reidenbach2023coarsenconf}.

\textbf{Computer Vision}

Computer vision entails the inference of properties of the visual world from images or other measurements such as LIDAR. There are many subtasks in computer vision, such as object recognition, semantic segmentation, image and video generation, and depth estimation. Historically, network primitives that capture the topological structure and symmetries of images have dominated vision benchmarks, including CNNs, GCNNs, and Vision Transformers (ViTs) \citep{LeCun1998Gradient-BasedRecognition, Krizhevsky2012ImageNetCW, cohen16, dosovitskiy2021image}. {\color{black}Within our taxonomy, CNNs operate on regular image grids and process Euclidean signals on Euclidean domains (Card S1, Figure~\ref{fig:data}). When translation equivariance is built in—as in standard CNNs—or extended to include rotational symmetries via group convolutions (e.g., GCNNs), these models align with Cards S7 and S10. ViTs similarly operate on Euclidean domains (Card S1) but with a different architectural inductive bias, typically treating images as sequences of patches with learned positional encodings. When symmetry-preserving mechanisms such as equivariant attention are introduced (see Figure~\ref{fig:alg_transformers}), ViTs may extend toward Card S10, provided both the positional structure and patch embeddings transform under a group action.}

Another successful application of non-Euclidean deep learning in computer vision has been graph neural networks. Specifically, these are implemented on data native to or lifted to point-clouds. Pioneering works such as Pointnet++ and PointTransformer introduced graph-structured deep networks as breakthrough methods in 3D semantic segmentation and object detection at whole-room scales \citep{qi2017pointnetpp, Engel2020PointT}. Another promising computational primitive, Slot Attention, introduces a novel messaging-passing strategy to perform unsupervised object discovery using permutation invariant slots, which incorporates spatial symmetries using slot-centric reference frames \citep{locatello2020objectcentric, biza2023invariant}.

\textbf{Biomedical Imaging}

Biomedical imaging involves inferring the structure of biological tissues from measurements of their physical properties, typically in the form of electromagnetic fields, acoustic waves, and other physical phenomena. Quantities of interest include shape, composition, or internal state. Thus, geometric and topological structure play an important role in their analysis. 

Many machine learning problems in medical imaging require reasoning about 3D structures, including their shape, their variations throughout a population, and changes throughout time. As tissue states are non-Euclidean, their statistics and evolution require geometric treatment \citep{Pennec_Sommer_Fletcher_2019}. Variations in organ shapes lie on low-dimensional manifolds, and geometry-aware dimensionality-reduction methods such as tangent PCA \citep{boisvert_geometric_2008} or Principle Geodesic Analysis (PGA) enable meaningful representations for downstream tasks \citep{fletcher2004principal, Fletcher_Joshi_2007, hinkle2012}. {\color{black}We refer to row 3 of Fig. \ref{fig:geo-dimension-reduction} for details on these techniques.} Geometric methods have also been applied the analysis of the effects of aging in the corpus collosum with MRI scans \citep{hinkle2012polynomial}, to brain connectomics data in Diffusion Tensor Imaging \citep{Pennec2006b}, and to the segmentation of 3D anatomical structures from CT and MRI scans in the lateral cerebral ventricle, the kidney parenchyma and pelvis, and the hippocampus \citep{Pizer2003}.

\textbf{Physics}

Physics data naturally has many symmetries and often takes the form of relations between unordered sets, an ideal setting for topological and equivariant methods. Dynamics between particles or nodes in a mesh can be effectively computed using learned graph message passing for various types of physics data \citep{sanchezgonzalez2020learning, pfaff2021learning}. Equivariant transformer {\color{black}(Fig. \ref{fig:alg_transformers}, row 1)} and graph neural network architectures {\color{black}(Fig. \ref{fig:topo_deep_learning}, row 4)} have been successfully applied to data analysis for the Large Hadron Collider and other simulations such as gravity for the n-body problem \citep{fuchs2020se3, brandstetter2022geometric}. Topological methods are well suited to process the hundreds of petabytes of highly relational data produced by experiments from the Large Hadron Collider and have demonstrated their utility for the next stage of fundamental discoveries in particle physics \citep{dezoort2023graph}. Recent work \citep{brehmer2023geometric} proposes an equivariant transformer architecture processing embedded geometric representations, and demonstrates its efficacy on mesh interaction estimation and n-body simulations. Astrophysics data is also well suited for application of equivariant networks. Some examples include the classification of radio galaxies using group-equivariant CNNs \citep{Scaife_2021}, optimal cosmological analysis using equivariant normalizing flows \citep{dai2022translation}, and cosmic microwave background radiation analysis using spherical equivariant CNNs \citep{mcewen2022scattering}.

\normalsize 

\begin{figure}
    \centering
    \includegraphics[width=\linewidth]{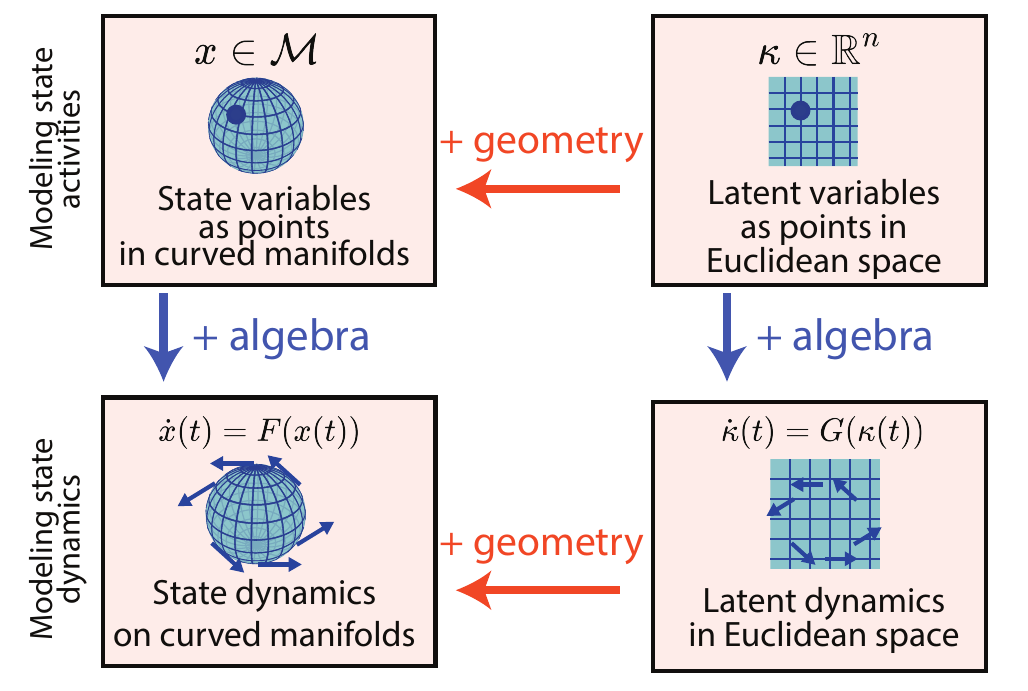}
    \caption{\textbf{Dynamics as algebra on state and latent variables}. In many applications, observed quantities evolve in time according to a low-dimensional latent dynamical systems. The arrows indicate algebraic (dynamics) or geometrical (structures representing real-world constraints) additions. Notation: $\mathbb{R}^n$: Euclidean space of dimension $n$, $M$: Manifold,  $x$: state variables as data points, $\kappa$: latent variables as data points, $F(\cdot)/G(\cdot)$: dynamical flow maps defined as algebras on Euclidean spaces and/or manifolds.}
    \label{fig:dynamics}
\end{figure}

{\color{black} 
\textbf{Outlook: Non-Euclidean Dynamical Systems}

Looking ahead, many scientific challenges, including those reviewed above, increasingly require not only understanding the potentially non-Euclidean structure of data, but also modeling how that structure evolves over time. From the folding and interaction of proteins, to the progression of disease in biomedical imaging, to predicting object trajectories and physical interactions in computer vision, the temporal evolution of complex systems is a central concern. While modeling static structure has led to significant improvements in their respective fields, as we reviewed above, incorporating geometry- or topology-constrained dynamics into machine learning is a natural and necessary next step.

Work in theoretical machine learning and dynamical systems theory, specifically on latent variable models, provides a promising foundation for addressing such problems (Fig.\ref{fig:dynamics}). Prior studies have shown that across diverse domains the observed variables often live in high-dimensional state spaces with geometric constraints (Fig.\ref{fig:dynamics}, \textit{left}). However, latent variables, whether in natural language processing \citep{mikolov2013efficient}, neuroscience \citep{schneider2023learnable,dinc2025latent}, or interpretability research \citep{shai2024transformers,hyvarinen2024identifiability}, can often be extracted with dimensional bottlenecks and consequently reside on low-dimensional manifolds, both Euclidean and non-Euclidean (Fig.~\ref{fig:dynamics}, \textit{right}). These latent spaces yield compact, interpretable, and physically meaningful representations of system dynamics.

A key opportunity for future work lies in modeling not just these latent states, but their evolution (Fig.~\ref{fig:dynamics}, \textit{bottom}). Many current models assume linear (Euclidean) dynamics \citep{abbaspourazad2024dynamical}, but manifold-valued dynamics can capture richer constraints like symmetry, curvature, and conservation laws. As in static settings, incorporating such geometric inductive biases may be crucial for generalization under limited data. Non-Euclidean dynamical systems thus offer a compelling direction for future models grounded in the structures introduced here.}

\vspace{-0.2cm}
\section{Conclusion}
\normalsize

As the availability of richly structured, non-Euclidean data grows, a new paradigm of machine learning has emerged, leveraging the mathematics of geometry, topology, and algebra to extract novel insights. In this review, we have provided an accessible overview of this field, unifying disparate threads in the literature into a common framework. Our illustrated taxonomy contextualizes, classifies, and differentiates existing approaches and illuminates gaps that present opportunities for innovation. In addition, we provide resources for the practitioner's use. Section \ref{sec:benchmarks} organizes the benchmarks used in the non-Euclidean deep learning literature. Section \ref{sec:software} provides a list of core open-source software libraries for non-Euclidean machine learning. Section \ref{sec:apps} summarizes the core domains that non-Euclidean machine learning has been applied to thus far.  We hope this serves as an invitation for both theoreticians and practitioners to further explore the potential for geometry, topology, and algebra to reshape modern machine learning, just as they reshaped our fundamental understanding of space over a century ago.

\section*{Acknowledgements}
 \footnotesize

M. P. acknowledges funding from the National Science Foundation (NSF) grant 2134241 and the Natural Sciences and Engineering Research Council of Canada; S. S. from the NSF grant 2313150; L. C. from the Chan Zuckerberg Initiative and the NSF Graduate Research Fellowship under grant 2139319; A.B. from the NIH R01NS119468 and the UCSB Chancellor's Fellowship; F. D. from grant NSF PHY-2309135 and the Gordon and Betty Moore Foundation Grant No. 2919.02 to the Kavli Institute for Theoretical Physics. X. P. was supported by ERC grant 786854 G-Statistics from the European Research Council under the European Union’s Horizon 2020 research and innovation program and by the French government through the 3IA Côte d’Azur Investments ANR-23-IACL-0001 managed by the National Research Agency. N. M. acknowledges funding from the NSF CAREER 2240158, NSF 2134241 and NSF 2313150.

\bibliography{bibliography}

\end{document}